\documentclass[journal,transmag]{IEEEtran}
\ifCLASSINFOpdf
\else
\fi
\usepackage{amsmath}
 \usepackage{graphicx}
  \usepackage{subfigure}
  \usepackage{epstopdf}
  \usepackage{hyperref}
\usepackage{cases}

\begin{document}
\title{Depth and Reflection Total Variation for Single Image Dehazing}


\author{\IEEEauthorblockN{Wei Wang\IEEEauthorrefmark{},
Chuanjiang He\IEEEauthorrefmark{}}
\IEEEauthorblockA{\IEEEauthorrefmark{}College of Mathematics and Statistics, Chongqing University, Chongqing 401331, China}% <-this % stops an unwanted space
\thanks{Manuscript received Jan, 2016; revised . This work was in part by the Chinese National Science Foundation (NSFC61561019, NSFC11371384) and Chongqing Graduate Student Research Innovation Project (CYS14020).
Corresponding author: Chuanjiang He (email: cjhe@cqu.edu.cn).}}

\markboth{ Jan~2016}%
{Shell \MakeLowercase{\textit{et al.}}: Bare Demo of IEEEtran.cls for IEEE Transactions on Magnetics Journals}



% for Transactions on Magnetics papers, we must declare the abstract and
% index terms PRIOR to the title within the \IEEEtitleabstractindextext
% IEEEtran command as these need to go into the title area created by
\IEEEtitleabstractindextext{%
\begin{abstract}
\emph{Abstract-} Haze removal has been a very challenging problem due to its ill-posedness, which is more ill-posed if the input data is only a single hazy image. In this paper, we present a new approach for removing haze from a single input image. The proposed method combines the model widely used to describe the formation of a haze image with the assumption in Retinex that an image is the product of the illumination and the reflection. We assume that the depth and reflection functions are spatially piecewise smooth in the model, where the total variation is used for the regularization. The proposed model is defined as a constrained optimization problem, which is solved by an alternating minimization scheme and the fast gradient projection algorithm. Some theoretic analyses are given for the proposed model and algorithm. Finally, numerical examples are presented to demonstrate that our method can restore vivid and contrastive hazy images effectively.
\end{abstract}

\begin{IEEEkeywords}
dehazing, Retinex, total variation, variational method, gradient projection algorithm.
\end{IEEEkeywords}}

\maketitle

\IEEEdisplaynontitleabstractindextext

% \fi
\IEEEpeerreviewmaketitle

\section{Introduction}
\IEEEPARstart{I}{n} foggy and hazy days, the light reflected from an object is usually absorbed and scattered by the turbid medium such as particles and water droplets in the atmosphere before it reaches the camera lens. Moreover, the light captured by the camera lens is often blended with the atmospheric light. This inevitably causes outdoor images taken in bad weather (e.g., foggy or hazy) to lose intensity contrast and color fidelity. Moreover, because most automatic systems strongly depend on the definition of the input images, they usually fail to work on the hazy images. Therefore, haze removal (or dehazing) is very important for the computer vision applications, such as remote sensing \cite{b1}, video surveillance systems \cite{b2}, and so on.

In computer vision and computer graphics, the model widely used to describe formation of a hazy image is that a hazy scene appearance is a weighted sum of the attenuated scene radiance and the transmitted atmospheric light whose extents depend on the scene distance to the camera lens (depth information) \cite{b31}. However, the depth information and the value of the atmospheric light are usually unknown in practice. Therefore, haze removal is a very challenging problem. The problem will be more ill-posed if the input data is only a single hazy image \cite{b15}.

In the literature, there exist mainly two classes of approaches for removing haze. A class of approaches is dependent on multiple images or additional information. Polarization methods \cite{b9}\cite{b10} remove haze by using the difference of the polarization characteristics from two or more images of the same scene taken with different polar angles. In \cite{b3}\cite{b11}\cite{b12}, depth-based methods are proposed, which obtain the depth information from the user inputs or from known 3D models. These methods are effective for image dehazing, but they are somewhat impractical because both multiple images and a 3D geometrical model are usually difficult to acquire in practice.

Another class of approaches is to only use a single image to remove haze. This class can also be divided into two categories: filter methods and energy methods. The filter methods benefit from new image models or stronger priors. Fattal \cite{b13} recovered the haze-free scene contrast, under the assumption that the optical transmission and surface shading are not locally correlated to the estimation of the albedo of the scene and the optical transmission. Though Fattal's assumption is physically reasonable, it is invalid when the haze is heavy or the assumption fails. Tan \cite{b14} removed haze by maximizing the local contrast of the restored image based on the observation that haze-free images have higher contrast than hazy images. Their results are greatly enhanced, but tend to be over-saturated, and the results may not be physically valid. Tarel and Hautière \cite{b19} introduced a fast algorithm to dehaze both color images and gray level images by using an edge preserving median filter. He et al. \cite{b15} proposed a dark channel prior (DCP) based on the observation that in most of the local regions it is very often that some pixels have very low intensity in at least one color (RGB) channel. Using this prior, they can recover high-quality haze-free images. However, this method has a high computational complexity in the soft matting process. After the DCP was proposed, the approaches based on this prior develop rapidly. In \cite{b16}, He et al. proposed a guided image filtering in place of soft matting to speed up the process of dehazing. In \cite{b18}, based on a physical model and the DCP, Wang et al. proposed a single image dehazing method which chooses the atmospheric light founded on a Variogram. Chen and Huang \cite{b24} presented a new image haze removal approach based on Fisher’s linear discriminant-based dual DCP scheme in order to solve the problems associated with the presence of localized light sources and color shifts, and thereby achieved effective restoration. In \cite{b27}, Choi et al. proposed a referenceless perceptual fog density prediction model based on natural scene statistics (NSS) and fog aware statistical features. This model only makes use of measurable deviations from statistical regularities observed in natural foggy and fog-free images, which not only assesses the performance of defogging algorithms designed to enhance the visibility of foggy images, but also is well suited for image defogging. In \cite{b23}, Huang et al. presented a novel Laplacian-based visibility restoration approach to solve inadequate haze thickness estimation and alleviate color cast problems. In \cite{b32}, Dai et al. proposed an adaptive sky detection and preservation method to dehaze the images. The method detects the sky by hard or smooth threshold of the sky area which selects the threshold value as a function of the histogram of the image to process.

The energy methods usually propose an energy functional that should be minimized or maximized and derive a numerical scheme. Nishino et al. \cite{b20} proposed a Bayesian probabilistic dehazing method that jointly estimates the scene albedo and depth from a single foggy image by fully leveraging their latent statistical structures. Their method is effective for dehazing images but needs to tune a lot of parameters that have great influence on the final results. In \cite{b17}, Fang et al. proposed a new energy model for dehazing and denoising simultaneously based on a windows adaptive DCP. In \cite{b21}, Fattal derived a local formation model that explains the color-lines which means pixels of small image patches typically exhibit a 1D distribution in RGB color space and used it for recovering the scene transmission based on the lines’ offset from the origin. In \cite{b7}, Wang et al. presented a multiscale depth fusion (MDF) method for defog from a single image. The fusion is formulated as an energy minimization problem that incorporates spatial Markov dependence. They got the depth map by minimizing the nonconvex potential in the random field which was solved by an alternate optimization algorithm. In \cite{b22}, Zhu et al. proposed a Color Attenuation Prior to dehaze images. They recovered the depth information via creating a linear model for modeling the scene depth whose parameters are learned by a supervised learning method. In \cite{b25}, Galdran et al. extended a perception-inspired variational framework for single image dehazing by substituting the mean of the clean image for the value of the gray-world hypothesis and adding a set of new terms to the energy functional for maximizing the interchannel contrast. In \cite{b28}, Lai et al. derived the optimal transmission map directly from the haze model under two scene priors including locally consistent scene radiance and context-aware scene transmission. They introduced theoretic and heuristic bounds of scene transmission and incorporated the two scene priors to formulate a constrained minimization problem which solved by quadratic programming.

In this paper, we propose a new energy approach for removing haze from a single input image. Our model is a constrained total variation model which is deduced by combining the model widely used to describe the formation of a haze image with the assumption in Retinex that an image is the product of the illumination and the reflection. Then we use an alternating minimization scheme and the fast gradient projection (FGP) algorithm \cite{b26} to solve the proposed model and give some theoretical analysis for our model and the algorithm. At last, some experimental results are presented to show the effectiveness of our method.

The rest of our paper is organized as follows. In Section 2, we deduce the proposed model and show the existence and uniqueness of solution for the model. In Section 3, we present the algorithm to solve our model and investigate its convergence. In Section 4, we give the experimental results and conclude in Section 5.

\section{ The proposed model}
In computer vision and computer graphics, the formulation widely used to describe the hazy image is as follows \cite{b31}:
\begin{equation}
   I(x) = t(x)J(x) + (1 - t(x))A,\quad x \in \Omega,
\end{equation}
where $I(x)$ is the observed intensity defined on the image domain $\Omega  \subset {R^2}$ , $J(x)$  is the radiance of the surface, $A$  is a constant representing the globally atmospheric light, and  $t(x)$ is the scene transmission describing the portion of the light that is not scattered and reaches the camera lens. When the atmosphere is homogenous, the transmission $t(x)$ can be formulated as \cite{b31}:
\begin{equation}
   t(x) = {e^{ - \beta d(x)}}, \quad x \in \Omega ,
\end{equation}
where $\beta $  is the positive scattering coefficient of the atmosphere and  $d(x)$ is the depth of scene which is also positive.

According to the assumption widely used in Retinex \cite{b4}\cite{b5}\cite{b6}\cite{b13}\cite{b21}, the surface radiance can be written as a product of the reflectance and the illumination:
 \begin{equation}
   J(x) = R(x)L(x),\quad x \in \Omega ,
\end{equation}
where $R(x)$  and $L(x)$  is the reflectance and the illumination, respectively. The reflectance $0 \le R(x) \le 1$  is related to the physical characteristics of the surface. For the following theoretic analysis, we assume that $0 \le R(x) < 1$  a.e. (almost everywhere). Based on the physical characteristics of the material object, the constraint $R(x) < 1$  a.e. is practical.

Substituting (3) into (1), we have
 \begin{equation}
   I(x) = t(x)R(x)L(x) + (1 - t(x))A,\quad x \in \Omega.
\end{equation}

Because most hazy images are landscapes of outdoor scenarios, we make the following assumption:
\begin{equation}
   L(x) \le A,\quad x \in \Omega.
\end{equation}

Since  $R(x) < 1$ a.e., by (4) and (5) we obtain
 \setlength{\arraycolsep}{0.0em}
\begin{eqnarray}
\begin{aligned}
   I(x)& = t(x)R(x)L(x) + A - t(x)A \\
   &< t(x)A + A - t(x)A \\
   &= A, \quad \text{a.e.} \;x \in \Omega.
   \end{aligned}
\end{eqnarray}
\setlength{\arraycolsep}{5pt}
Equation (6) implies
 $$A \ge ess{\sup _{x \in \Omega }}I(x).$$
Because the atmospheric light  $A$ is a constant, we have
$$A = ess{\sup _{x \in \Omega }}I(x) + C_0,$$
where $C_0 \ge 0$  is a constant. By the way, by experiments we found that the proposed model is insensitive to the value of $C_0$ (or $A$), the assumption (5) is valid by choosing larger value of $C_0$.

Equation (4) can be rewritten as
 \begin{equation}
   1 - \frac{{I(x)}}{A} = t(x)(1 - \frac{{R(x)L(x)}}{A}),\quad x \in \Omega.
\end{equation}
Let $R^{*}(x)=\frac{{R(x)L(x)}}{A}$. Because of (5), we have $0\le R^{*} \le R< 1$. In particular, if $L(x) = A$, then $R^{*} = R$.
In order to handle the product expression in (7), we convert the product to the logarithmic form. Let
 $i(x) = \log (1 - \frac{{I(x)}}{A}),\quad\eta (x) = \log (t(x)) =  - \beta d(x), \quad\gamma (x) = \log(1 - R^{*}(x)). $
Then, Equation (7) is equivalent to
$$i(x) = \eta (x) + \gamma (x),\quad x \in \Omega.$$
Besides, since $- \beta d(x)<0$ and $0<1 - R^{*}(x)<1$, we have $\eta (x) \le 0$, $\gamma (x) \le 0$ and so $i(x) \le \eta (x)$, $i(x) \le \gamma (x)$. Therefore, the constraints $i \le \eta  \le 0$  and  $i \le \gamma  \le 0$, a.e. need to be added into the proposed model.

It should be pointed out that Equation (7) has the similar form with Equation (8) in \cite{b20}, but the deriving processes are different. We derive Equation (7) by merging the Retinex theory into the haze imaging model (1), while Nishino et al. \cite{b20} derive their model by rewriting the Koschmieder’s law. The constraint $\gamma (x) \le 0$ is very important for our variational method (see Figure 10). However, this constraint cannot be obtained by Nishino et al.' method of deriving Equation (8) in \cite{b20}.

We further make the following assumptions before presenting our model:
\begin{itemize}
  \item The depth function $\eta (x)$  and the reflection function $\gamma (x)$  are piecewise smooth in $\Omega $. Thus $\eta ,\;\gamma  \in BV(\Omega )$  and the regularization terms are defined by $\int_\Omega  {\left| {\nabla \eta } \right|} $  and  $\int_\Omega  {\left| {\nabla \gamma } \right|} $.
  \item The sum of the depth function $\eta (x)$  and the reflection function $\gamma (x)$  is close
to  $i(x)$. Thus, the fidelity term is given by  $\int_\Omega  {{{(\eta  + \gamma  - i)}^2}} $.
\end{itemize}

Collecting the above two assumptions into one expression, we propose the following energy functional for dehazing:
\begin{equation}
  E(\eta ,\gamma ) = \alpha \int_\Omega  {\left| {\nabla \eta } \right|}  + \int_\Omega  {{{(\eta  + \gamma  - i)}^2}}  + \beta \int_\Omega  {\left| {\nabla \gamma } \right|}+\lambda \int_\Omega{\gamma^2},
\end{equation}
where  $\alpha>0$ and $\beta>0$  are regularization parameters and $\lambda>0$ is an arbitrary small constant.

The term $\lambda \int_\Omega{\gamma^2}$ is only used for the proof of the uniqueness of the minimizer. Without this term, a constant can be added to variable $\eta$ and subtracted to variable $\gamma$ without a change of the value of the energy functional. Numerically, there is no difference for our model with $\lambda$ being a very small value (such as $10^{-10}$) and $\lambda$ being zero. Therefore, we always set $\lambda = 0$ in numerical implementation.

 The proposed model is thus given by the constrained optimization problem:
\begin{equation}
  \begin{aligned}
  &\text{Minimize:}\quad\; E(\eta ,\gamma ),\;\;\;\eta ,\;\gamma  \in BV(\Omega )\\
  &\text{Subject to:}\quad i \le \eta  \le 0\;,\;i \le \gamma  \le 0,\;\text{a.e.}
  \end{aligned}
\end{equation}

We first show the existence of solution for the above optimization problem.

\newtheorem{Theorem}{Theorem}
\begin{Theorem}
The functional $E(\eta ,\gamma )$ is strictly convex in $\left\{ {(\eta ,\;\gamma )|\;i \le \eta ,\gamma  \le 0} \right\}$.
\end{Theorem}

The proof is trivial and so is omitted.

\begin{Theorem}
If  $i \in {L^2}(\Omega )$, then problem (9) has exactly one solution.
\end{Theorem}

\begin{IEEEproof} First, the energy functional $E(\eta ,\gamma )$  is clearly nonnegative and proper because $E(0,0)$  is a finite value. Let $({\eta _n},{\gamma _n})$  is a minimizing sequence of problem (9), i.e.,
 $$\mathop {\lim }\limits_{n \to \infty } E({\eta _n},{\gamma _n}) = \inf_{{\eta ,\gamma  \in BV(\Omega ),i \le \eta  \le 0,i \le \gamma  \le 0}}E(\eta ,\gamma ).$$
Then, there is a constant $C>0$  so that
\begin{equation*}
\begin{aligned}
 E({\eta _n},{\gamma _n}) =&\alpha \int_\Omega  {\left| {\nabla {\eta _n}} \right|}  + \int_\Omega  {{{({\eta _n} + {\gamma _n} - i)}^2}}  + \beta \int_\Omega  {\left| {\nabla {\gamma _n}} \right|}\\
 &+\lambda \int_\Omega{\gamma_n^2}  \le C.
 \end{aligned}
 \end{equation*}
Thus,
\begin{equation}
   \int_\Omega  {\left| {\nabla {\eta _n}} \right|}  \le C/\alpha,\;\;\int_\Omega  {\left| {\nabla {\gamma _n}} \right|}  \le C/\beta.
\end{equation}

Since  $i \in {L^2}(\Omega )$, $i \le {\eta _n} \le 0$  and  $i \le {\gamma _n} \le 0$, we have
 $$\int_\Omega  {\eta _n^2}  \le \int_\Omega  {{i^2}},\int_\Omega  {\gamma _n^2}  \le \int_\Omega  {{i^2}}.$$
By the Jensen's inequality, we obtain
\begin{equation}
\begin{aligned}
   &\int_\Omega  {\left| {{\eta _n}} \right|}  \le {\left| \Omega  \right|^{1/2}}{\left\| {{\eta _n}} \right\|_{{L^2}(\Omega )}} \le {\left| \Omega  \right|^{1/2}}{\left\| i \right\|_{{L^2}(\Omega )}},\\
   &\int_\Omega  {\left| {{\gamma _n}} \right|}  \le {\left| \Omega  \right|^{1/2}}{\left\| i \right\|_{{L^2}(\Omega )}}.
\end{aligned}
\end{equation}
Namely, the sequences $\{ {\eta _n}\} $  and $\{ {\gamma _n}\} $  are bounded in  ${L^1}(\Omega )$. Combining it with (10), we know that the sequences $\{ {\eta _n}\} $  and  $\{ {\gamma _n}\} $ are bounded in  $BV(\Omega )$. Hence, there exists some $({\eta _*},{\gamma _*}) \in BV(\Omega ) \times BV(\Omega )$  such that, up to a subsequence,
   \begin{equation}
    {\eta _n} \to {\eta _*}\;\;\text{and}\;\;{\gamma _n} \to {\gamma _*}\;\;\text{in}\;{L^1}(\Omega ),
   \end{equation}
\begin{equation}
    {\eta _n} \to {\eta _*}\;\;\text{and}\;\;{\gamma _n} \to {\gamma _*}\;\;\text{weakly in}\;{L^2}(\Omega ).
   \end{equation}
By the (weak) lower semicontinuity of the  ${L^2}-\text{norm}$ and (13), we have
\begin{equation}
 \int_\Omega  {{{({\eta _*} + {\gamma _*} - i)}^2}}  \le \mathop {\liminf}\limits_{n \to \infty } \int_\Omega  {{{({\eta _n} + {\gamma _n} - i)}^2}}.
\end{equation}
By (12) and the lower semicontinuity of  $BV(\Omega )$, we have
\begin{equation}
\begin{aligned}
    \int_\Omega  {\left| {\nabla {\eta _*}} \right|}  \le \mathop {\liminf}\limits_{n \to \infty } \int_\Omega  {\left| {\nabla {\eta _n}} \right|}\\
    \int_\Omega  {\left| {\nabla {\gamma _*}} \right|}  \le \mathop {\liminf}\limits_{n \to \infty } \int_\Omega  {\left| {\nabla {\gamma _n}} \right|}.
\end{aligned}
\end{equation}
Recalling the semicontinuity of  $L^2$ norm, by (13) we have
\begin{equation}
  \int_\Omega{\gamma_*^2}  \le \mathop {\liminf}\limits_{n \to \infty } \int_\Omega  {\gamma_n^2}.
\end{equation}

Combining (14) and (15), we further have
\begin{equation*}
  \begin{aligned}
&E({\eta _*},{\gamma _*})\\
=&\alpha \int_\Omega  {\left| {\nabla {\eta _*}} \right|}  + \int_\Omega  {{{({\eta _*} + {\gamma _*} - i)}^2}}  + \beta \int_\Omega  {\left| {\nabla {\gamma _*}} \right|}+ \lambda \int_\Omega{\gamma_*^2}\\
\le&\mathop {\alpha \liminf}\limits_{n \to \infty } \int_\Omega  {\left| {\nabla {\eta _n}} \right|}  + \mathop {\liminf}\limits_{n \to \infty } \int_\Omega  {{{({\eta _n} + {\gamma _n} - i)}^2}}  \\
&+\mathop { \beta\liminf}\limits_{n \to \infty } \int_\Omega  {\left| {\nabla {\gamma _n}} \right|}+\lambda \mathop {\liminf}\limits_{n \to \infty } \int_\Omega  {\gamma_n^2} \\
\le& \mathop {\liminf}\limits_{n \to \infty } (\alpha \int_\Omega  {\left| {\nabla {\eta _n}} \right|}+\int_\Omega  {{{({\eta _n}+{\gamma _n} - i)}^2}}+\beta \int_\Omega {\left| {\nabla {\gamma _n}} \right|}+\lambda\int_\Omega{\gamma_n^2}) \\
=& \mathop {\liminf}\limits_{n \to \infty } E({\eta _n},{\gamma _n})\\
=& \mathop {\lim }\limits_{n \to \infty } E({\eta _n},{\gamma _n})\\
=& \mathop {\inf }\limits_{\eta ,\gamma  \in BV(\Omega ),i \le \eta  \le 0,i \le \gamma  \le 0} E(\eta ,\gamma ).
\end{aligned}
\end{equation*}

Meanwhile, by (12) we have, up to a subsequence,
 \begin{equation}
 {\eta _n} \to {\eta _*}\quad \text{and}\quad {\gamma _n} \to {\gamma _*}\quad\text{a.e. in} \;\Omega .
 \end{equation}
By $i \le {\eta _n} \le 0$  and $i \le {\gamma _n} \le 0$  a.e., we obtain that $({\eta _*},{\gamma _*})$  satisfies the constraints $i \le {\eta _*} \le 0$  and  $i \le {\gamma _*} \le 0$, a.e.. Thus  $({\eta _*},{\gamma _*})$ is a solution of problem (9).
The uniqueness of the solution is guaranteed by the strictly convex of the functional $E(\eta ,\gamma )$.
\end{IEEEproof}

\section{The algorithm}

Since there are two unknown variables in problem (9), we use the alternating minimization scheme to convert it into two subproblems. We describe it in Algorithm 1 in detail.\\

\textbf{Algorithm 1.}\\

\begin{itemize}
  \item Set initial value ${\eta _0} = i$,  ${\gamma _0} = 0$,  $k = 0$, predefined iterations ${N_1} > 0$  and error tolerance  $\varepsilon >0$.
  \item At the $k$th iteration:\\
      \begin{enumerate}
        \item Given  ${\gamma _k}$, compute ${\eta _{k + 1}}$  by solving\\
        \begin{equation}
           \mathop {\min }\limits_{i \le \eta  \le 0} {E_1}(\eta ) = \alpha \int_\Omega  {\left| {\nabla \eta } \right|}  + \int_\Omega  {{{(\eta  + {\gamma _k} - i)}^2}}.
        \end{equation}
        \item Given  ${\eta _{k + 1}}$, compute ${\gamma _{k + 1}}$  by solving\\
        \begin{equation}
        \begin{aligned}
           &\mathop {\min }\limits_{i \le \gamma  \le 0} {E_2}(r) = \\
           &\beta \int_\Omega  {\left| {\nabla \gamma } \right|}  + \int_\Omega  {{{(\gamma  + {\eta _{k + 1}} - i)}^2}}+\lambda \int_\Omega{\gamma^2}.
        \end{aligned}
        \end{equation}
      \end{enumerate}
  \item Go back to step 2 until $\frac{{\left\| {{\eta _{k + 1}} - {\eta _k}} \right\|}}{{\left\| {{\eta _{k + 1}}} \right\|}} \le \varepsilon \;\& \; \frac{{\left\| {{\gamma _{k + 1}} - {\gamma _k}} \right\|}}{{\left\| {{\gamma _{k + 1}}} \right\|}} \le \varepsilon $ or $k = {N_1}$.
\end{itemize}\

Problems (18) and (19) are the constrained total variation minimal problems, which can be solved by the FGP algorithm \cite{b26} efficiently.

Suppose that the hazy image $I(x)$  is an $m \times n$  matrix. Let  $p \in {R^{(m - 1) \times n}}$, $q \in {R^{m \times (n - 1)}}$  and  $r \in {R^{m \times n}}$, define:
\begin{itemize}
  \item The unit ball in ${R^{(m - 1) \times n}} \times {R^{m \times (n - 1)}}$:
  $$E = \left\{ {(p,q)\left| {\begin{array}{*{20}{c}}
{p_{i,j}^2 + q_{i,j}^2 \le 1,\;\;{\begin{array}{*{20}{c}}
{i = 1,...,m - 1,}\\
{j = 1,...,n - 1}
\end{array}}}\\
{\left| {{p_{i,n}}} \right| \le 1,\;\;\;\;\;\;\;\;\;\;\;i = 1,...,m - 1}\\
{\left| {{q_{n,j}}} \right| \le 1,\;\;\;\;\;\;\;\;\;\;\;j = 1,...,n - 1}
\end{array}} \right.} \right\}$$
  \item The divergence of $\left( {p,q} \right)$:
  \begin{align*}
    &{\left( {{\mathop{\rm div}\nolimits} \left( {p,q} \right)} \right)_{i,j}} =\\
     &\left\{ \begin{array}{l}
{p_{i,j}} + {q_{i,j}} - {p_{i,j}} - {q_{i,j}},\;{\begin{array}{*{20}{c}}
{i = 1,...,m - 1,}\\
{j = 1,...,n - 1}
\end{array}}\\
0,\;\;\;\;\;\;\;\;\;\;\;\;\;\;\;\;\;\;\;\;\;\;\;\;\;\;\;i = 1,m,j = 1,...,n\\
0,\;\;\;\;\;\;\;\;\;\;\;\;\;\;\;\;\;\;\;\;\;\;\;\;\;\;\;i = 1,...,m,j = 1,n
\end{array} \right.
  \end{align*}
  \item The gradient of $r$:
  $$\nabla (r) = (p,q),$$
  where $$p = {r_{i,j}} - {r_{i + 1,j}},\;\;i = 1,...,m - 1,j = 1,...,n$$
  $$q = {r_{i,j}} - {r_{i,j + 1}},\;\;i = 1,...,m,j = 1,...,n - 1$$
  \item The projection from $r$ on $C = [i,\;0]$:
  $${P_C}(r) = \min (\max (i,r),0)$$
  \item  The projection from $(p,q)$ on $E$:
  $${\left( {{P_E}(p,q)} \right)_{(i,j)}} = \left( {\frac{{{p_{i,j}}}}{{\max (1,\left| {{p_{i,j}}} \right|)}},\frac{{{q_{i,j}}}}{{\max (1,\left| {{q_{i,j}}} \right|)}}} \right)$$
\end{itemize}

Then, according to Proposition 4.1 in \cite{b26}, the solution ${\eta _{k + 1}}$  of problem (18) can be written as
$${\eta _{k + 1}} = {P_C}\left[ {i - {\gamma _k} - \alpha {\mathop{\rm div}\nolimits} (p,q)} \right],$$
where $(p,q)$  is the solution of the following problem:
\begin{equation}
\begin{aligned}
   \mathop{\min}\limits_{(p,q) \in E}  h(p,q)= &- {{\left\| {{H_C}(i - {\gamma _k} - \alpha {\mathop{\rm div}\nolimits} (p,q))} \right\|}^2}\\
    &+ {{\left\| {i - {\gamma _k} - \alpha {\mathop{\rm div}\nolimits} (p,q)} \right\|}^2}
\end{aligned}
\end{equation}
with  ${H_C}(r) = r - {P_C}(r)$. Problem (20) is computed by
\begin{equation}
   ({p_k},{q_k}) = {P_E}( {({l_k},{s_k}) + \frac{1}{{8\alpha }}\nabla ({P_C}\left[ {i - {\gamma _k} - \alpha {\mathop{\rm div}\nolimits} ({l_k},{s_k})} \right])} ),
\end{equation}
where $({l_k},{s_k})$  is obtained by an accelerated scheme from  $({p_{k - 1}},{q_{k - 1}})$:
\begin{equation}
\begin{aligned}
    &{t_{k + 1}} = \frac{{1 + \sqrt {1 + 4t_k^2} }}{2},\\
    &({l_{k + 1}},{s_{k + 1}}) = ({p_k},{q_k}) + \frac{{{t_k} - 1}}{{{t_{k + 1}}}}({p_k} - {p_{k - 1}},{q_k} - {q_{k - 1}}).
\end{aligned}
\end{equation}
Thus, the FGP algorithm for Problem (18) is as follows:\\

\textbf{Algorithm 2.}\\

\begin{itemize}
  \item Set initial values $({l_1},{s_1}) = ({p_0},{q_0}) = 0$, ${t_1} = 1,$ $j = 1$  and predefined iterations  ${N_2} > 0$.
  \item At the jth iteration:\\
  Given $({p_{j - 1}},{q_{j - 1}})$  and  $({l_j},{s_j})$, Compute
  \begin{enumerate}
    \item $({p_j},{q_j})$ by (21),
    \item ${t_{j + 1}}$  and $({l_{j + 1}},{s_{j + 1}})$  by (22).
  \end{enumerate}
  \item  $j = j + 1$, go back to step 2 until convergence or  $j = {N_2}$.
  \item Set  $output = {P_C}\left[ {i - {\gamma _j} - \alpha {\mathop{\rm div}\nolimits} ({p_j},{q_j})} \right]$.
\end{itemize}\

Since we set $\lambda=0$, we can solve problem (19) by the similar method.

Next, we investigate the convergence of Algorithm 1.

\begin{Theorem}
 Let $i \in {L^2}(\Omega )$  and $({\eta _k},{\gamma _k})$  be the sequence derived from Algorithm 1. Then the sequence $({\eta _k},{\gamma _k})$  converges to some $({\eta _\# },{\gamma _\# }) \in BV(\Omega ) \times BV(\Omega )$ (up to a sequence), and for any
 $$(\eta ,\gamma ) \in \left\{ {(\eta ,\gamma )|\eta ,\gamma  \in BV(\Omega ),i \le \eta  \le 0,i \le \gamma  \le 0,a.e.} \right\},$$
we have
$$E({\eta _\# },{\gamma _\# }) \le E({\eta _\# },\gamma ),\quad E({\eta _\# },{\gamma _\# }) \le E(\eta ,{\gamma _\# }).$$
\end{Theorem}

\begin{IEEEproof}
 By (18) and (19), we have the following inequalities
$$E({\eta _{k + 1}},{\gamma _{k + 1}}) \le E({\eta _{k + 1}},{\gamma _k}) \le E({\eta _k},{\gamma _k}) \le ... \le E({\eta _0},{\gamma _0}).$$
Thus, $E({\eta _k},{\gamma _k})$  is bounded and decreases with  $k$. Because of
\begin{equation}
 \begin{aligned}
 E({\eta _k},{\gamma _k})=& \alpha \int_\Omega  {\left| {\nabla {\eta _k}} \right|}+\int_\Omega  {{{({\eta _k} + {\gamma _k} - i)}^2}}\\
 &+ \beta\int_\Omega  {\left| {\nabla {\gamma _k}} \right|}+\lambda \int_\Omega{\gamma_k^2}.\\
\le& E({\eta _0},{\gamma _0}),
 \end{aligned}
\end{equation}

we have
\begin{equation}
\int_\Omega  {\left| {\nabla {\eta _k}} \right|}  \le E({\eta _0},{\gamma _0})/\alpha, \quad \int_\Omega  {\left| {\nabla {\gamma _k}} \right|}  \le E({\eta _0},{\gamma _0})/\beta.
\end{equation}

Since  $i \in {L^2}(\Omega )$, $i \le {\eta _k} \le 0$  and  $i \le {\gamma _k} \le 0$, we have
 $$\int_\Omega  {\eta _k^2}  \le \int_\Omega  {{i^2}},\quad \int_\Omega  {\gamma _k^2}  \le \int_\Omega  {{i^2}},$$
which implies that $\{ {\eta _k}\} $  and $\{ {\gamma _k}\} $  are bounded in ${L^2}(\Omega )$  and so bounded in  ${L^1}(\Omega )$. By (24), we obtain that $\{ {\eta _k}\} $  and $\{ {\gamma _k}\} $  are bounded in  $BV(\Omega )$. Therefore, up to a subsequence, there is some $({\eta _\# },{\gamma _\# }) \in BV(\Omega ) \times BV(\Omega )$ so that
\begin{equation}
 {\eta _k} \to {\eta _\# },\; {\gamma _k} \to {\gamma _\# } \;\text{in} \; L^1(\Omega ),
\end{equation}

\begin{equation}
 {\eta _k} \to {\eta _\# },\; {\gamma _k} \to {\gamma _\# } \;\text{weakly in} \; L^2(\Omega ).
\end{equation}

Since  $E({\eta _k},{\gamma _k}) \ge 0$ and decreases with  $k$, there exists $M \ge 0$  such that
$$\mathop {\lim }\limits_{k \to \infty } E({\eta _k},{\gamma _k}) = M.$$

Now, we prove  $E({\eta _\# },{\gamma _\# }) = M$.

In fact, by the lower semicontinuity of the  ${L^2}-\text{norm} $ and  $BV(\Omega )$, (25) and (26), we have
\begin{equation}
   \begin{aligned}
&\int_\Omega  {{{({\eta _\# } + {\gamma _\# } - i)}^2}}  \le \mathop {\liminf}\limits_{k \to \infty } \int_\Omega  {{{({\eta _k} + {\gamma _k} - i)}^2}},\\
&\int_\Omega  {{{\gamma _\#}^2}}  \le \mathop {\liminf}\limits_{k \to \infty } \int_\Omega  {{{{\gamma _k}}^2}},\\
 &\int_\Omega  {\left| {\nabla {\eta _\# }} \right|}  \le \mathop {\liminf}\limits_{k \to \infty } \int_\Omega  {\left| {\nabla {\eta _k}} \right|},\\
 &\int_\Omega  {\left| {\nabla {\gamma _\# }} \right|}  \le \mathop {\liminf}\limits_{k \to \infty } \int_\Omega  {\left| {\nabla {\gamma _k}} \right|}.
 \end{aligned}
   \end{equation}
Therefore, we further have
   \begin{equation}
   \begin{aligned}
M\; =&\mathop {\liminf}\limits_{k \to \infty } E({\eta _k},{\gamma _k})\\
\ge& \alpha \mathop {\liminf}\limits_{k \to \infty } \int_\Omega  {\left| {\nabla {\eta _k}} \right|}  + \mathop {\liminf}\limits_{k \to \infty } \int_\Omega  {{{({\eta _k} + {\gamma _k} - i)}^2}}  \\
&+ \beta \mathop {\liminf}\limits_{k \to \infty } \int_\Omega  {\left| {\nabla {\gamma _k}} \right|}+\lambda \mathop {\liminf}\limits_{k \to \infty } \int_\Omega  {{{{\gamma _k}}^2}}\\
\ge& \alpha \int_\Omega  {\left| {\nabla {\eta _\# }} \right|}  + \int_\Omega  {{{({\eta _\# } + {\gamma _\# } - i)}^2}}\\
&+ \beta \int_\Omega  {\left| {\nabla {\gamma _\# }} \right|} +\lambda\int_\Omega  {{{\gamma _\#}^2}}\\
=&E({\eta _\# },{\gamma _\# }).
\end{aligned}
   \end{equation}

On the other hand, by considering the inequalities
$$E({\eta _{k + 1}},{\gamma _{k + 1}}) \le E({\eta _{k + 1}},{\gamma _k}) \le E({\eta _\# },{\gamma _k})$$
and
 $$E({\eta _{k + 1}},{\gamma _{k + 1}}) \le E({\eta _k},{\gamma _k}) \le E({\eta _k},{\gamma _\# }),$$
we have
\begin{equation}
  2E({\eta _{k + 1}},{\gamma _{k + 1}}) \le E({\eta _\# },{\gamma _k}) + E({\eta _k},{\gamma _\# }).
\end{equation}

The right-hand side of (29) can explicitly be written as
\begin{equation*}
   \begin{aligned}
&E({\eta _\# },{\gamma _k}) + E({\eta _k},{\gamma _\# }) \\
= &\alpha \int_\Omega  {\left| {\nabla {\eta _\# }} \right|}  + \int_\Omega  {{{({\eta _\# } + {\gamma _k} - i)}^2}}  + \beta \int_\Omega  {\left| {\nabla {\gamma _k}} \right|}+\lambda\int_\Omega  {{{\gamma _k}^2}} \\
& + \alpha \int_\Omega  {\left| {\nabla {\eta _k}} \right|}  + \int_\Omega  {{{({\eta _k} + {\gamma _\# } - i)}^2}}  + \beta \int_\Omega  {\left| {\nabla {\gamma _\# }} \right|}+\lambda\int_\Omega  {{{\gamma _\#}^2}}.
\end{aligned}
   \end{equation*}
Since
\begin{equation*}
   \begin{aligned}
 &\int_\Omega  {{{({\eta _\# } + {\gamma _k} - i)}^2}}  + \int_\Omega  {{{({\eta _k} + {\gamma _\# } - i)}^2}} \\
 = &\int_\Omega  {{{({\eta _\# } + {\gamma _\# } - i)}^2}}  + \int_\Omega  {{{({\eta _k} + {\gamma _k} - i)}^2}} \\
 &+ 2\int_\Omega  {({\eta _k} - {\eta _\# })({\gamma _k} + {\gamma _\# })},
 \end{aligned}
   \end{equation*}
by (29) we have
\begin{equation*}
   \begin{aligned}
2E({\eta _{k + 1}},{\gamma _{k + 1}}) \le& E({\eta _\# },{\gamma _k}) + E({\eta _k},{\gamma _\# }) \\
\le& E({\eta _\# },{\gamma _\# }) + E({\eta _k},{\gamma _k}) \\
&+ 2\int_\Omega  {({\eta _k} - {\eta _\# })({\gamma _k} - {\gamma _\# })}.
\end{aligned}
   \end{equation*}
Letting  $k \to \infty $, we have
$$2M \le E({\eta _\# },{\gamma _\# }) + M,$$
which implies  $E({\eta _\# },{\gamma _\# }) \ge M$. Combining (28), we obtain  $E({\eta _\# },{\gamma _\# }) = M$.

Next, we show the second assertion.

In fact, for any  $\eta  \in BV(\Omega )$,  $i \le \eta  \le 0$, we have
$$E({\eta _{k + 1}},{\gamma _{k + 1}}) \le E({\eta _{k + 1}},{\gamma _k}) \le E(\eta ,{\gamma _k}),$$
$$E({\eta _{k + 1}},{\gamma _{k + 1}}) \le E({\eta _k},{\gamma _k}) \le E({\eta _k},{\gamma _\# }).$$
The above two inequalities imply that
$$2E({\eta _{k + 1}},{\gamma _{k + 1}}) \le E({\eta _k},{\gamma _k}) + E(\eta ,{\gamma _\# }) + 2\int_\Omega  {({\eta _k} - \eta )({\gamma _k} - {\gamma _\# })}. $$
Letting  $k \to \infty $, we have $$E({\eta _\# },{\gamma _\# }) \le E(\eta ,{\gamma _\# }).$$
Similarly, for any  $r \in BV(\Omega )$,  $i \le r \le 0$, we have
$$E({\eta _\# },{\gamma _\# }) \le E({\eta _\# },\gamma )$$
The proof is completed.
\end{IEEEproof}

\section{Experimental results}
In this section, we present some experimental results to show the performance and the effectiveness of our model, in comparison to other relevant models such as TH model \cite{b19}, HST model \cite{b15}, NN model \cite{b20} and GVPB model \cite{b25}.

For the proposed model, we set $\alpha  = 100$ , $\beta  = 0.1$ and $\lambda=0$ ; ${N_1} = 100$  and $\varepsilon  = 0.1$  for Algorithm 1 and ${N_2} = 100$ for Algorithm 2. For the other comparison models excluding  NN model, we use the default parameters
provided in their corresponding papers (or programs). For NN model, we use the airlight $A=[0.7, 0.7, 0.7]$ in the program downloaded from the authors' homepage.

For color images, we process each of three color channels (R,G,B) separately and then synthesize the final results. For each color channel, the atmospheric light is theoretically estimated by:
 $${A_c} = {\max _{x \in \Omega }}{I_c}(x)+C_0,$$
where $c \in \left\{ {R,G,B} \right\}$  is color channel index, and $C_0$ is a positive constant. Since our model is insensitive to the value of $C_0$ (Figure 6 gives an example), $A_c$ is simply set as 255 in all experiments. The transmission is obtained by
 $${t_c}(x) = {{\mathop{\rm e}\nolimits} ^{{\eta _c}(x)}},$$
where ${\eta _c}(x)$  is derived from Algorithm 1. With the transmission ${t_c}(x)$  and the atmospheric light  ${A_c}$, we can theoretically recover the each color channel of the scene radiance according to Equation (1); namely,
\begin{equation}
    {J_c}(x) = \frac{{{I_c}(x) - {A_c}}}{{{t_c}(x)}} + {A_c} \\
\end{equation}
However, since the transmission ${t_c}(x)$  may be very close to zero in some patches (e.g., sky regions), it is known from (30) that the directly recovered scene radiance will contain some black patches. As done in \cite{b15}, we restrict the transmission to a lower bound  ${t_0} > 0$. Therefore, the scene radiance is recovered by
\begin{equation}
    {J_c}(x) = \frac{{{I_c}(x) - {A_c}}}{{\max ({t_c}(x),{t_0})}} + {A_c},\;c \in \left\{ {R,G,B} \right\}.
\end{equation}
Besides, since the scene radiance obtained by (31) looks a little dark, we perform a gamma correction with $gamma = 0.7$  before recovering the final scene radiance. The value of  ${t_0}$  can typically range from 0.1 to 0.4 by experiments. In the following experiments, we simply set ${t_0=0.4}$.

In the following, we present some experiments of our model, in comparison to other four models. The original hazy images are shown in Figure 1 (a) to Figure 5 (a). Figure 1(b-e) to Figure 5 (b-e) show the results of the TH model  \cite{b19}, the HST model \cite{b15}, the NN model \cite{b20} and the GVPB model \cite{b25}, respectively. The results of our model are given in Figure 1(f) to Figure 6(f). We see from Figure 1(b) to Figure 5 (b) that the haze of the results obtained by the TH model \cite{b19} is removed clearly and the detail in the pictures is well preserved. However, the results significantly suffer from over-enhancement. This effect can be easily observed in Figure 2(b) and Figure 4(b). Moreover, halo artifacts appear near the discontinuities (see the intersection areas of the trees and the sky in Figure 4(b)). This is because the TH model tends to produce a dark sky. From Figure 1(c) to Figure 5 (c) we can see that the HST model \cite{b15} produces natural results and can dehaze image effectively. But the detail of the results is not well (see the grass ground in Figure 1(c) and the rock in Figure 4(c)). Because the refined transmission of the HST model contains too much detail, it can't represent the depth information accurately. Besides, a halo effect can be observed in the sky in Figure 3(c). We can observe from Figure 1 (d) to Figure 5 (d) that NN model \cite{b20} is very effective for some images such as Figure 1 (d) and Figure 4(d). But it also suffers from over-enhancement for other images such as Figure 2 (d) Figure 4(d). Beside, we can see in Figure 5 (d), the result tends to be yellow. The reason why some of the results are good and others are not very satisfactory is that we set the parameters uniformly and the NN model is sensitive to the values of the parameters. We observe from Figure 1(e) to Figure 5(e) that the GVPB model \cite{b25} produces more natural results. But since it dehazes image based on enhancement which don't utilize the depth information, it can't clearly dehaze the far away region of the image whose scene depth varies from far to near. This effect can be easily observed in Figure 1(e) and Figure 4(e). Compared with the results of the four models, our results are free from halo effect and the details of our results are enhanced moderately, as displayed in Figure 1(f) to Figure 5(f). The sky in Figure 2(f) and Figure 3(f) is brighter and clear and the details in Figure 1(f) and Figure 4(f) are enhanced well. From Figure 5, we can see the result obtained by our method in  Figure 5 has the best visual effect.

In the next experiment, we test the sensitivity of the parameters $C_0$, $\alpha$, and $\beta$. First, we fix $\alpha=100$, $\beta=0.1$, and set $C_0=0, 10, 20$ and  30. From Figure 6(b)-Figure 6(e), we can observe that the change of the value of $C_0$ has almost no influence on the final results. Next, we fix $A=255$,  $\beta=0.1$ and set $\alpha=10, 100, 1000$. From Figure 7(b)-Figure 7(d), we can see that there is almost no change in the enhanced images when the parameter $\alpha$ varies from 10 to 1000. Last, we fix $A=255$,  $\alpha=100$ and set $\beta=0.01, 0.1, 1$. We can see from Figure 8(b)-Figure 8(d) that the results changes a little dark as the parameter $\beta$ varies from 0.01 to 1.

In this set of experiments, we evaluate the four comparison methods and our method quantitatively.  For this purpose, we need the ground truth images. Therefore, we download the fog-free images from the dataset \cite{b33} and use the graduated fog filter of the Photoshop to synthesize 48 hazy images. Figure 9 shows some of the results. We use the total  mean square error (MSE) index to evaluate these results objectively. The MSE is defined as:
$$MSE=\sqrt {\frac{\sum_{i=1}^3{(I_i-G_i)^2}}{3*m*n}},$$
where $I_i$ and $G_i$ are the obtained result and the ground truth, respectively, the subscript  $i \in \left\{ {R,G,B} \right\}$  is color channel index, and $m*n$ is the number of the image pixels. From the definition, we know that the smaller the value of the MSE is, the better the result is.
Table 1 lists the MSE values of the 48 images, from which we can see our method has the smallest MSE value averagely.

In the next experiment, we give a example to demonstrate the importance of the constraints $i \le \eta ,\gamma  \le 0$ in our model. Figure 10(a)-Figure 10 (c) are the hazy image and the dehazed images by our model without or with constraint, respectively. We can observe from Figure 10(b) that the model without the constraint can't dehaze the image at all.

Since our method tackles the image channel separately, it also works for the gray-scale images. Figure 11 shows an example.

\section{Conclusion and Discussion}
In this paper, we have introduced a new variational method for single image dehazing. The proposed method converts the problem of estimating the transmission map to a constrained variational problem. The existence and uniqueness of the solution of the proposed model are discussed. To solve the model efficiently, we adopt an alternating minimization scheme and the FGP algorithm. At last, we present some experimental results to test the effectiveness of our model.

It is still a challenging problem to dehaze image, especially for the images with both fine details and sky. Our method is to some extend effective for these images, but it still needs to further improve. For example, we simply choose $t_0=0.4$ in our method. This can make the results avoid the halo effect, but it is at the cost of losing some contrast. In the future, we will consider how to choose $t_0$ adaptively.

% or


%\appendices

%\section{}

%\section*{Acknowledgment}

\ifCLASSOPTIONcaptionsoff
  \newpage
\fi



%\bibliographystyle{IEEEtran}

\begin{thebibliography}{1}

\bibitem{b26}
{\sc A. Beck and M. Teboulle},
{\em Fast gradient-based algorithms for constrained total variation image denoising and deblurring problems}, IEEE Trans. Image Process., 18(2009), pp. 2419-2434.

\bibitem{b24}
{\sc B. Chen and S. Huang},
{\em An advanced visibility restoration algorithm for single hazy images}, ACM Trans. Multimed. Comput. Commun. Appl., 11(2015).

\bibitem{b27}
{\sc L. K. Choi, J. You, and A. C. Bovik},
{\em Referenceless prediction of perceptual fog density and perceptual image defogging}, IEEE Trans. Image Process., 24(2015), pp. 3888-3901.

\bibitem{b13}
{\sc R. Fattal},
{\em Single image dehazing}, In SIGGRAPH, 2008, pp. 1-9.

\bibitem{b17}
{\sc F. Fang, F. Li, and T. Zeng},
{\em Single image dehazing and denoising: A fast variational approach}, SIAM J. Imaging Sci., 7 (2014), pp. 969-996.


\bibitem{b21}
{\sc R. FATTAL},
{\em Dehazing Using Color-Lines}, ACM Trans. Graph., 34(2014), pp. 13:1-13:14

\bibitem{b25}
{\sc A. Galdran, J. Vazquez-Corral, D. Pardo, and M. Bertalm},
{\em Enhanced variational image dehazing}, SIAM J. Imaging Sci., 8(2015), pp. 1519-1546.

\bibitem{b15}
{\sc K. He, J. Sun, and X. Tang},
{\em Single image haze removal using dark channel prior}, IEEE Trans. Pattern Anal. Mach. Intell., 33 (2011), pp. 2341-2353.

\bibitem{b16}
{\sc K. He, J. Sun, and X. Tang},
{\em Guided image filtering}, IEEE Trans. Pattern Anal. Mach. Intell., 35 (2013), pp. 1397-1409.

\bibitem{b23}
{\sc S. Huang, J. Ye, and B. Chen},
{\em An advanced single-image visibility restoration algorithm for real-world hazy scenes}, IEEE Trans. Ind. Electron, 62(2015), pp. 2962-2972.

\bibitem{b4}
{\sc R. KIMMEL},
{\em A variational framework for Retinex}, Int. J. Comput. Vis., 52(2003), pp. 7-23.

\bibitem{b11}
{\sc J. Kopf, B. Neubert, B. Chen, M. Cohen, D. Cohen-Or, O. Deussen, M. Uyttendaele, and D. Lischinski},
{\em Deep photo: Model-based photograph enhancement and viewing}, ACM Trans. Graph., 27(2008).

\bibitem{b1}
{\sc J. Long, Z. Shi, W. Tang, and C. Zhang},
{\em Single remote sensing image dehazing}, IEEE Geosci. Remote Sensing Lett., 11 (2014), pp. 59-63.

\bibitem{b28}
{\sc Y. Lai, Y. Chen, C. Chiou, and C. Hsu},
{\em Single-image dehazing via optimal transmission map under scene priors}, IEEE Trans. Circuits Syst. Video Technol., 25(2015), pp. 1-14.

\bibitem{b8}
{\sc S. G. Narasimhan and S. K. Nayar},
{\em Vision and the atmosphere}, Int. J. Comput. Vis., 48(2002), pp. 233-254.

\bibitem{b12}
{\sc S. G. Narasimhan and S. K. Nayar},
{\em Interactive deweathering of an image using physical models}, In Workshop on Color and Photometric Methods in Computer Vision, 2003.

\bibitem{b3}
{\sc S. Narasimhan and S. Nayar},
{\em Contrast restoration of weather degraded images}, IEEE Trans. Pattern Anal. Mach. Intell., 25 (2003), pp. 713-724.

\bibitem{b5}
{\sc M. K. Ng and W. Wang},
{\em A total variation model for Retinex}, SIAM J. Imaging Sci., 4 (2011), pp. 345-365.

\bibitem{b20}
{\sc K. Nishino, L. Kratz and S. Lombardi},
{\em Bayesian defogging}, Int J Comput Vis, 98 (2012) pp. 263-278.

\bibitem{b9}
{\sc Y. Y. Schechner, S. G. Narasimhan, and S. K. Nayar},
{\em Instant dehazing of images using polarization}, CVPR, 1(2001), pp. 325-332.

\bibitem{b10}
{\sc S. Shwartz, E. Namer, and Y. Y. Schechner},
{\em Blind haze separation}, CVPR, 2(2006), pp. 1984 - 1991.

\bibitem{b14}
{\sc R. Tan},
{\em Visibility in bad weather from a single image}, CVPR, 2008, pp. 1-8.

\bibitem{b19}
{\sc J.-P. Tarel and N. Hautière},
{\em Fast visibility restoration from a single color or gray level image}, in Proc. IEEE Int. Conf. Comput. Vis., 2009, pp. 2201-2208.

\bibitem{b7}
{\sc Y. Wang and C. Fan},
{\em Single image defogging by multiscale depth fusion}, IEEE Trans. Image Process., 23 (2014), pp. 4826-4837.

\bibitem{b6}
{\sc W. Wang and C. He},
{\em A variational model with barrier functionals for Retinex}, SIAM J. Imaging Sci., 8 (2015), pp. 1955-1980.

\bibitem{b18}
{\sc  J. Wang, N. He, L. Zhang, and K. Lu},
{\em Single image dehazing with a physical model and dark channel prior}, Neurocomputing, 149 (2015), pp. 718-728.

\bibitem{b2}
{\sc I. Yoon, S. Kim, D. Kim, M.H. Hayes, and J. Paik},
{\em Adaptive defogging with color correction in the HSV color space for consumer surveillance system},IEEE Trans. Consum. Electron., 58 (2012), pp. 111-116.

\bibitem{b22}
{\sc Q. Zhu, J. Mai, and L. Shao},
{\em A fast single image haze removal algorithm using color attenuation prior}, IEEE Trans. Image Process., 24(2015), pp. 3522-3533.

\bibitem{b31}
{\sc Middleton},
{\em Vision through the atmosphere}, University of Toronto Press, 1952.

\bibitem{b32}
{\sc SK. Dai, JP, Tarel},
{\em Adaptive sky detection and preservation in dehazing algorithm}, Proc. IEEE International Symposium on Intelligent Signal Processing and Communication Systems, 2015.

\bibitem{b33}
{\em Natural fog-free, foggy, and test images used in FADE:}, \url{http://live.ece.utexas.edu/research/fog/fade_defade.html}.

\end{thebibliography}


\begin{IEEEbiography}{Wei Wang}
Biography text here.
\end{IEEEbiography}

\begin{IEEEbiography}{Chuanjiang He}
Biography text here.
\end{IEEEbiography}

%\newpage

%\vfill

\clearpage
\newpage

\begin{figure*}
\begin{center}
\subfigure[Hazy image]{\includegraphics[scale=0.3]{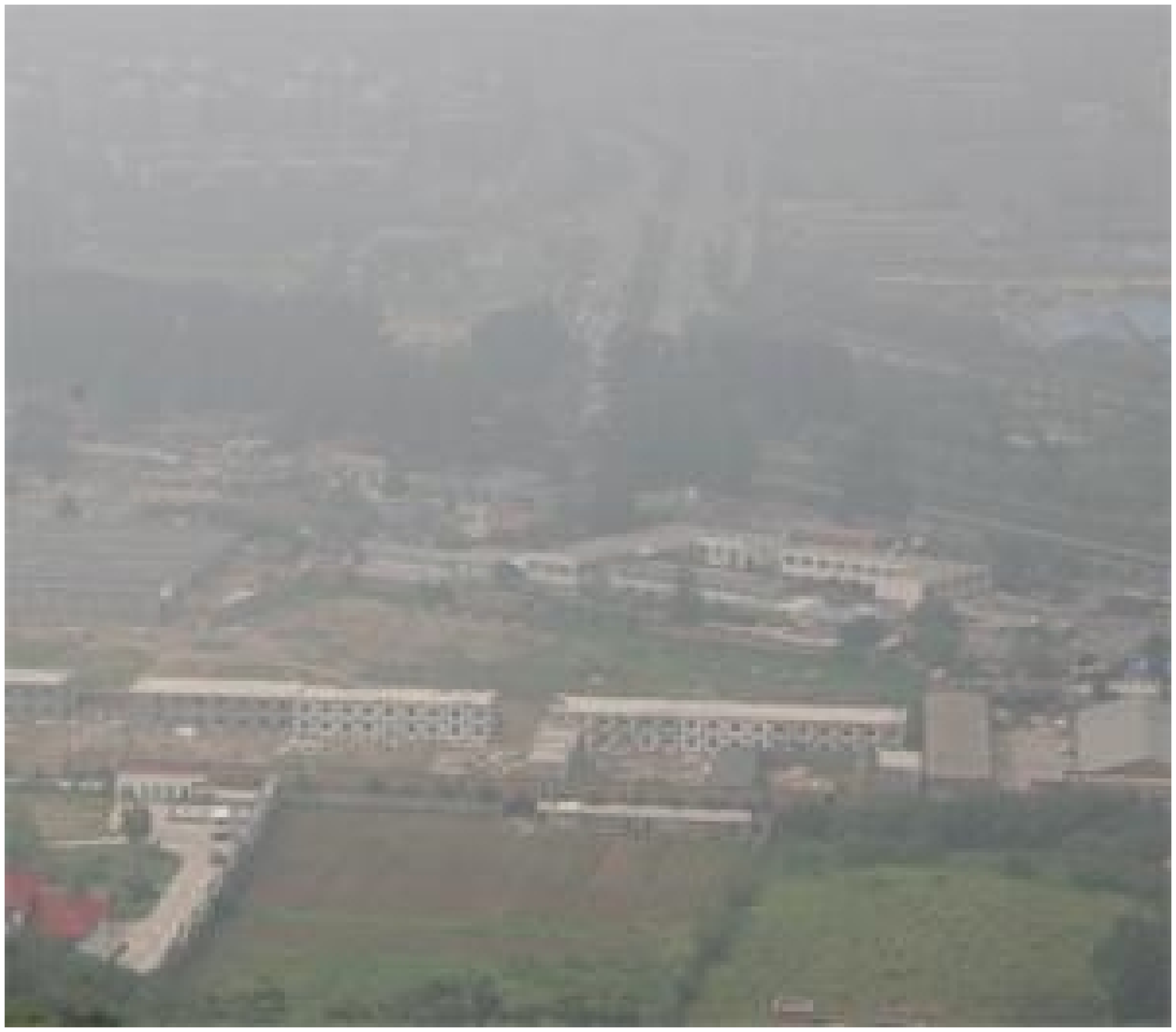}}
\subfigure[TH model]{\includegraphics[scale=0.3]{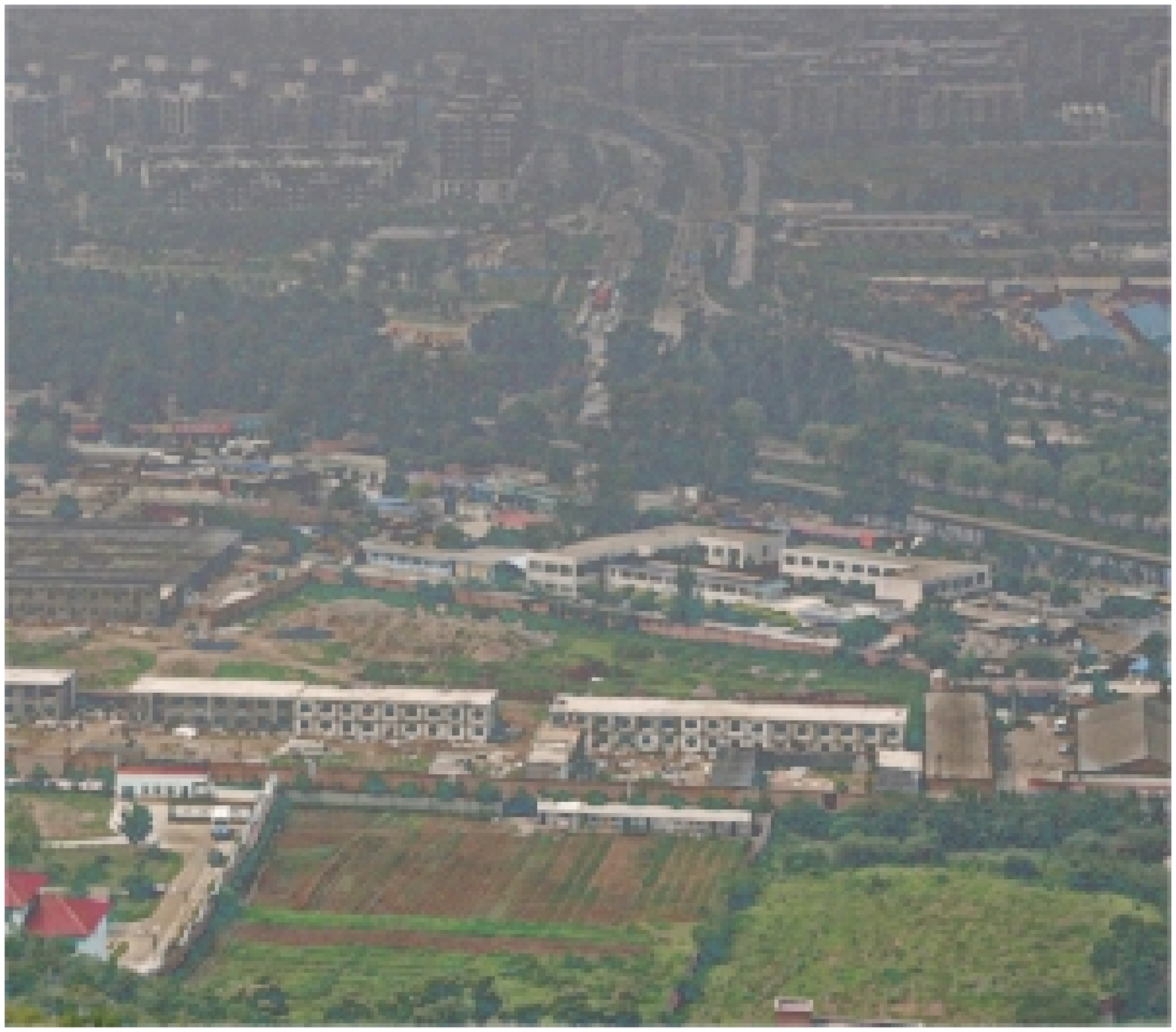}}
\subfigure[HST model]{\includegraphics[scale=0.3]{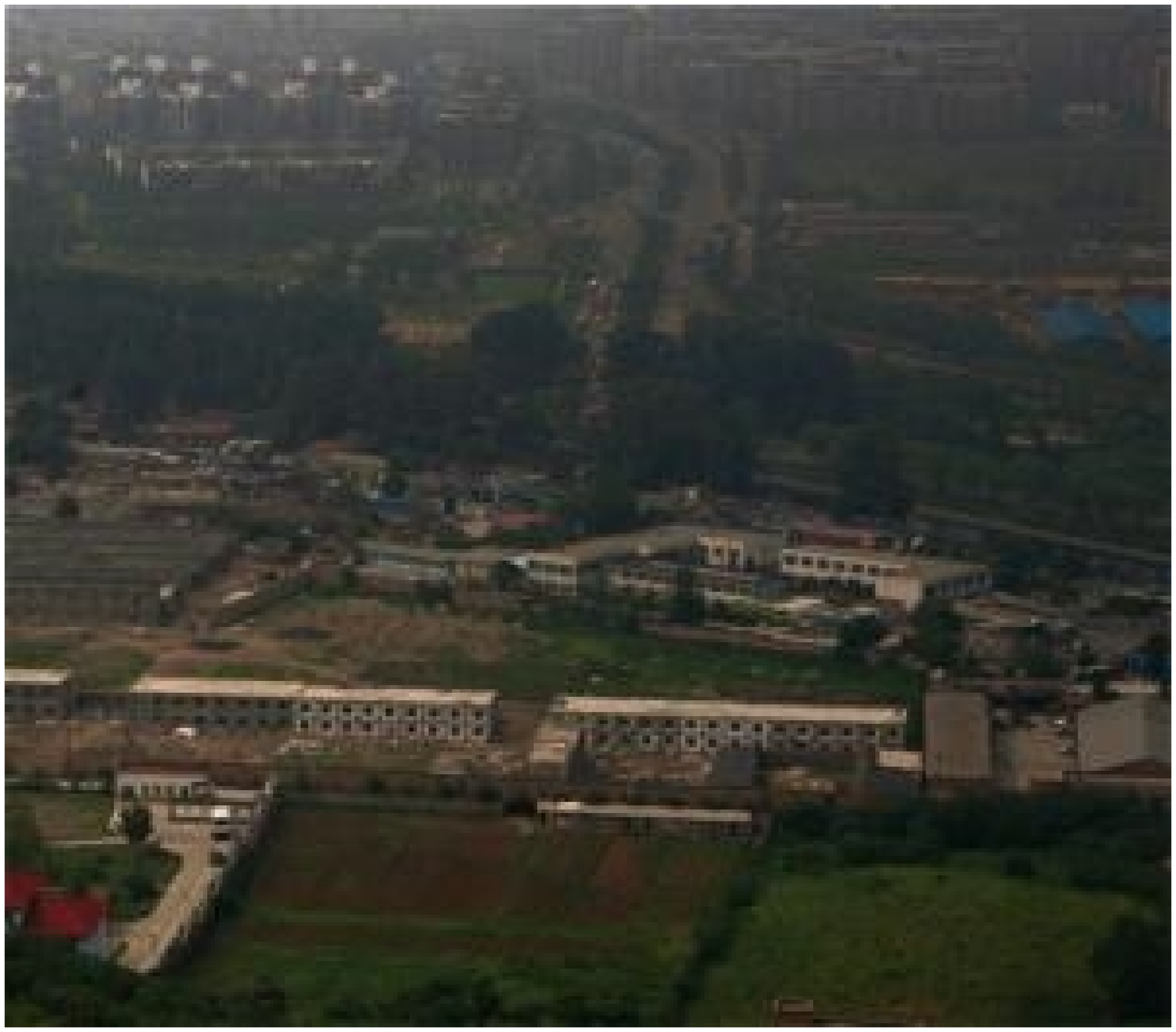}}
\subfigure[NN model]{\includegraphics[scale=0.3]{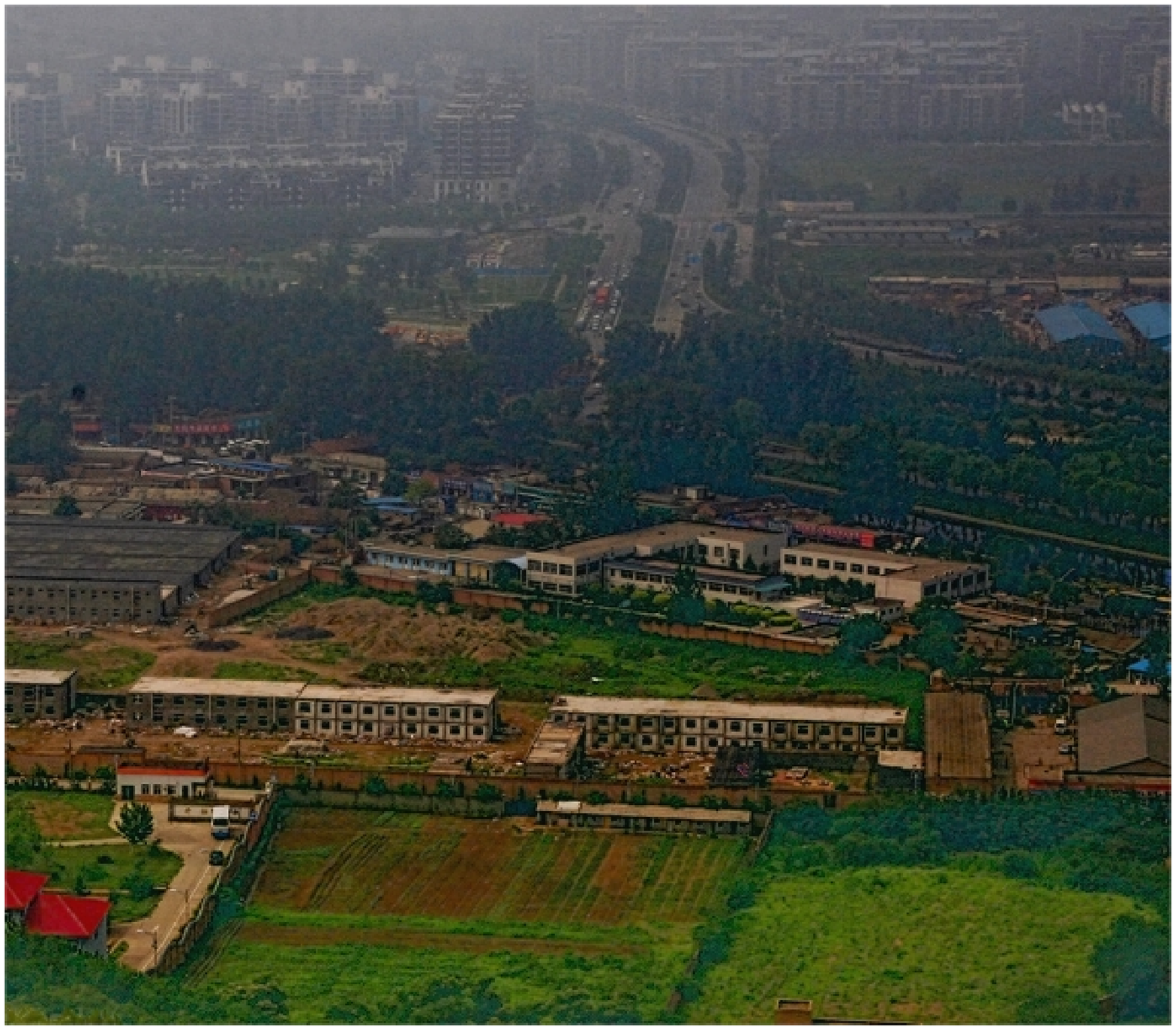}}
\subfigure[GVPB model]{\includegraphics[scale=0.3]{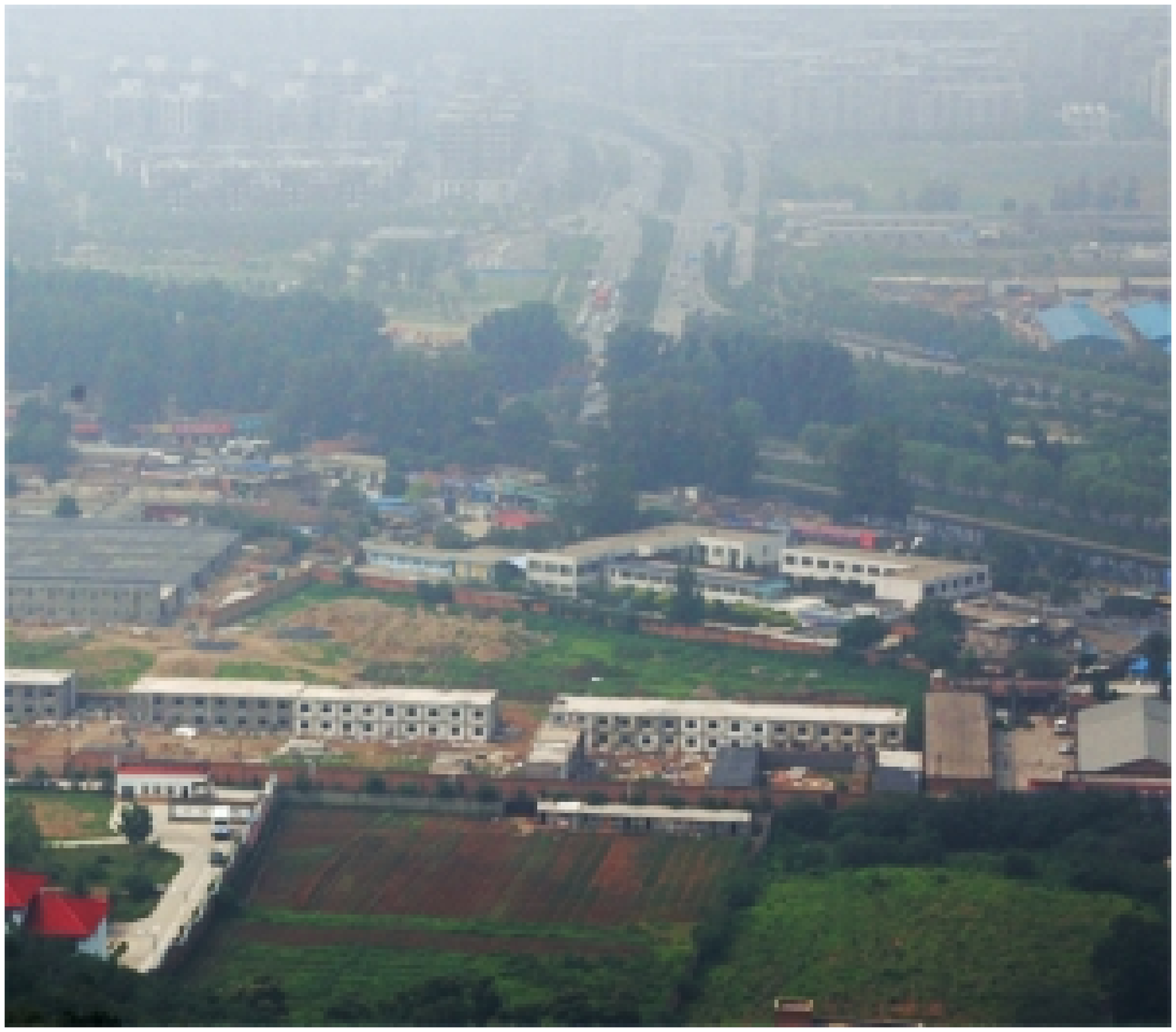}}
\subfigure[Our model]{\includegraphics[scale=0.3]{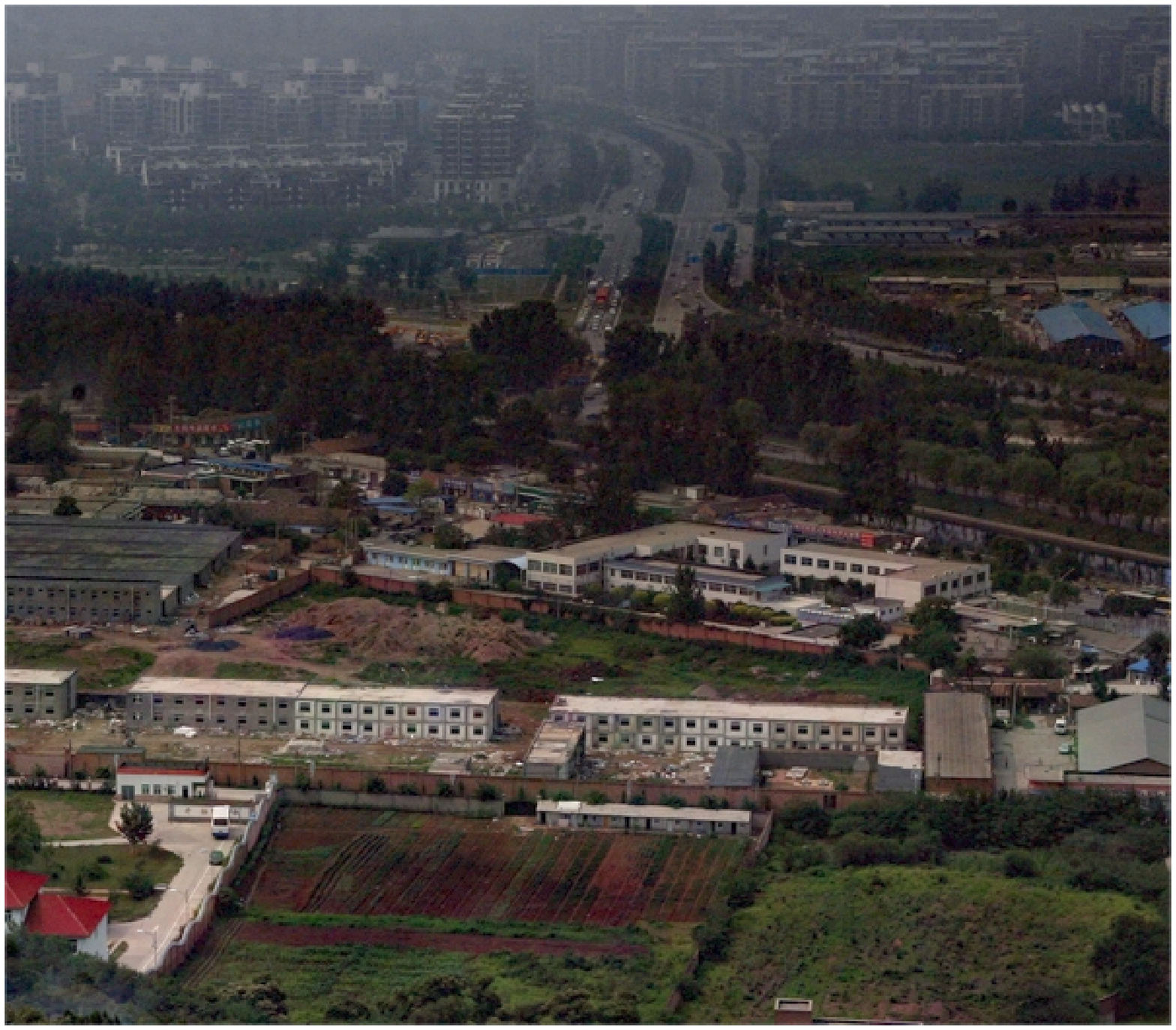}}
\end{center}
\caption{Dehazed images of five models.}
\end{figure*}

\begin{figure*}
\begin{center}
\subfigure[Hazy image]{\includegraphics[scale=0.26]{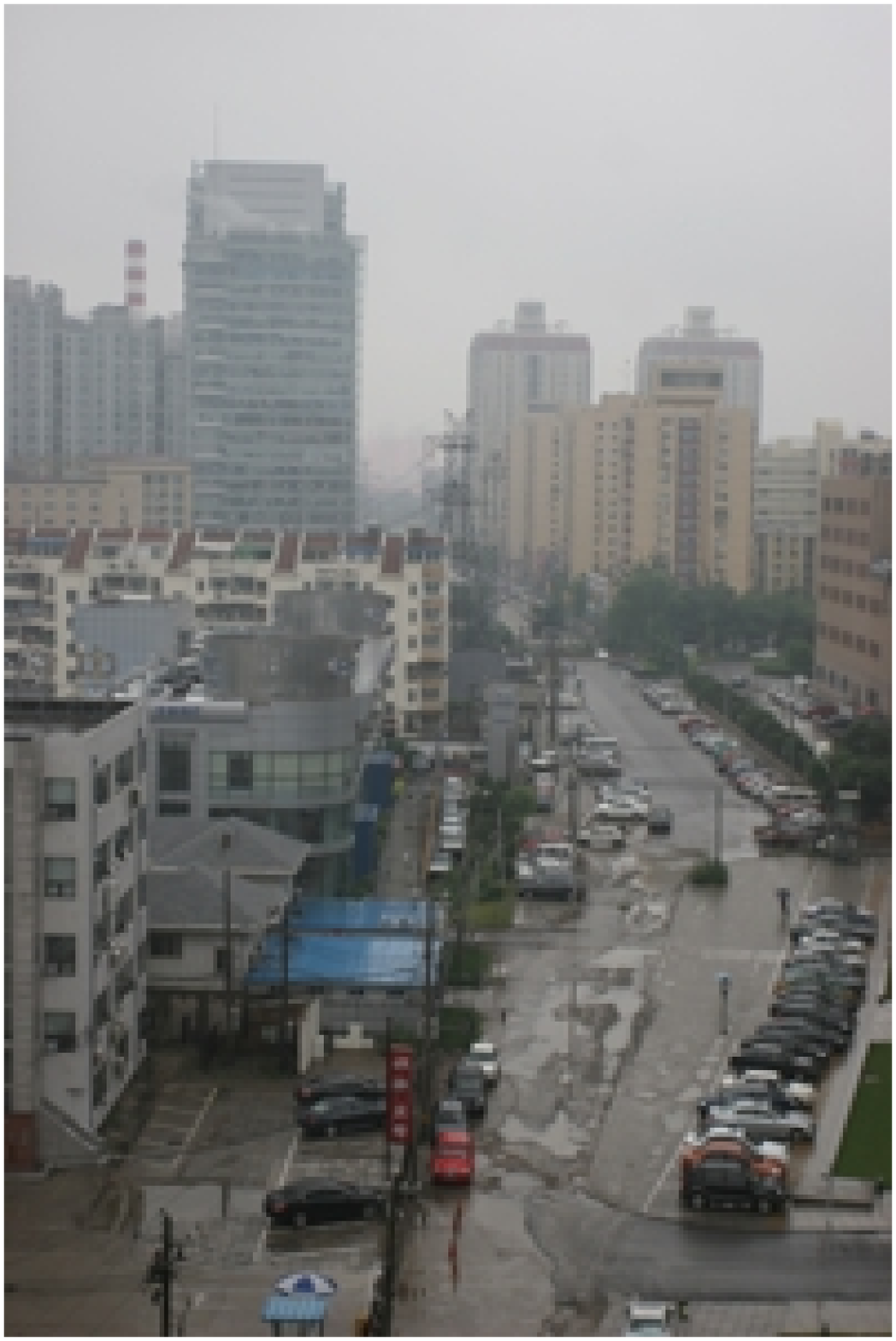}}
\subfigure[TH model]{\includegraphics[scale=0.26]{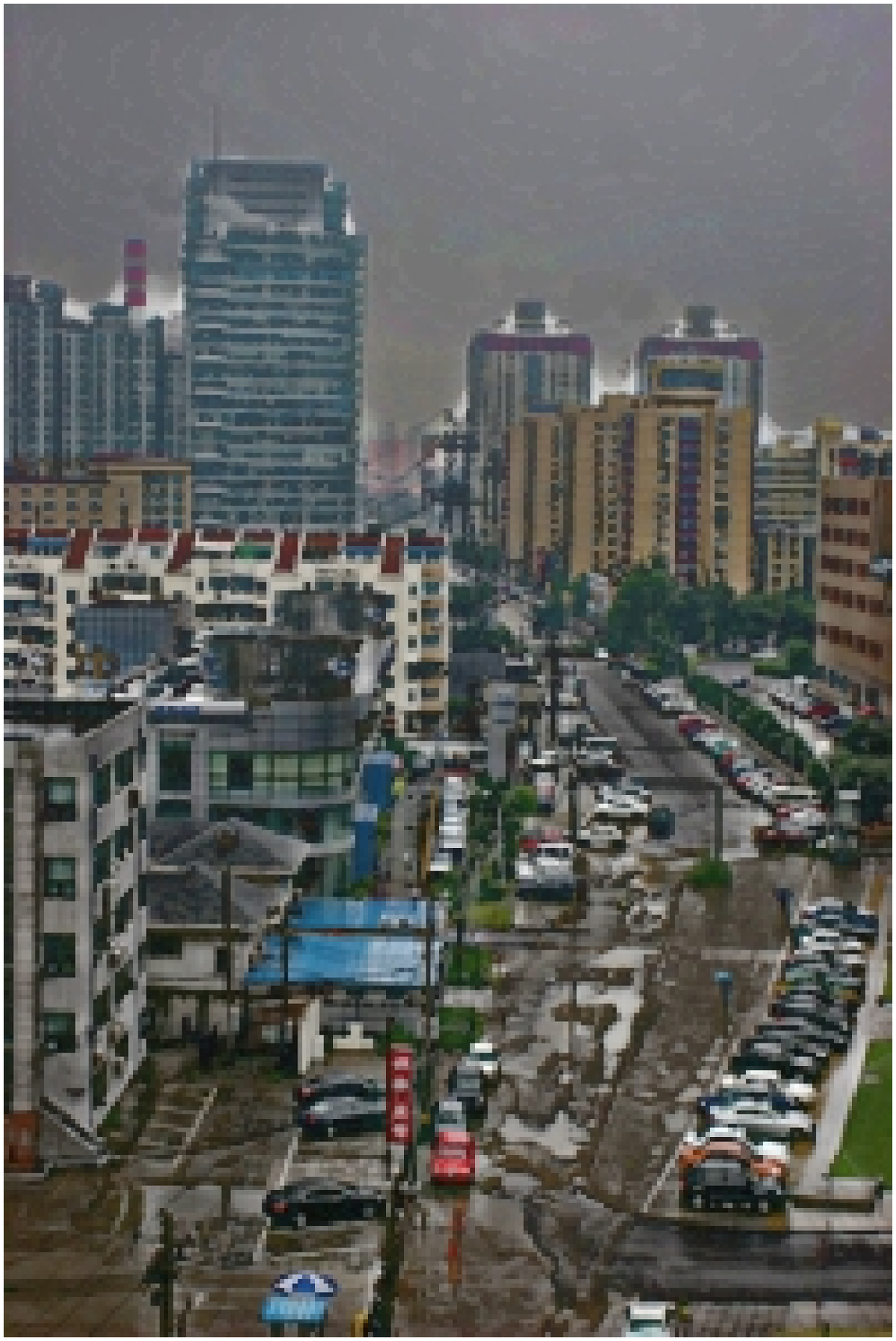}}
\subfigure[HST model]{\includegraphics[scale=0.26]{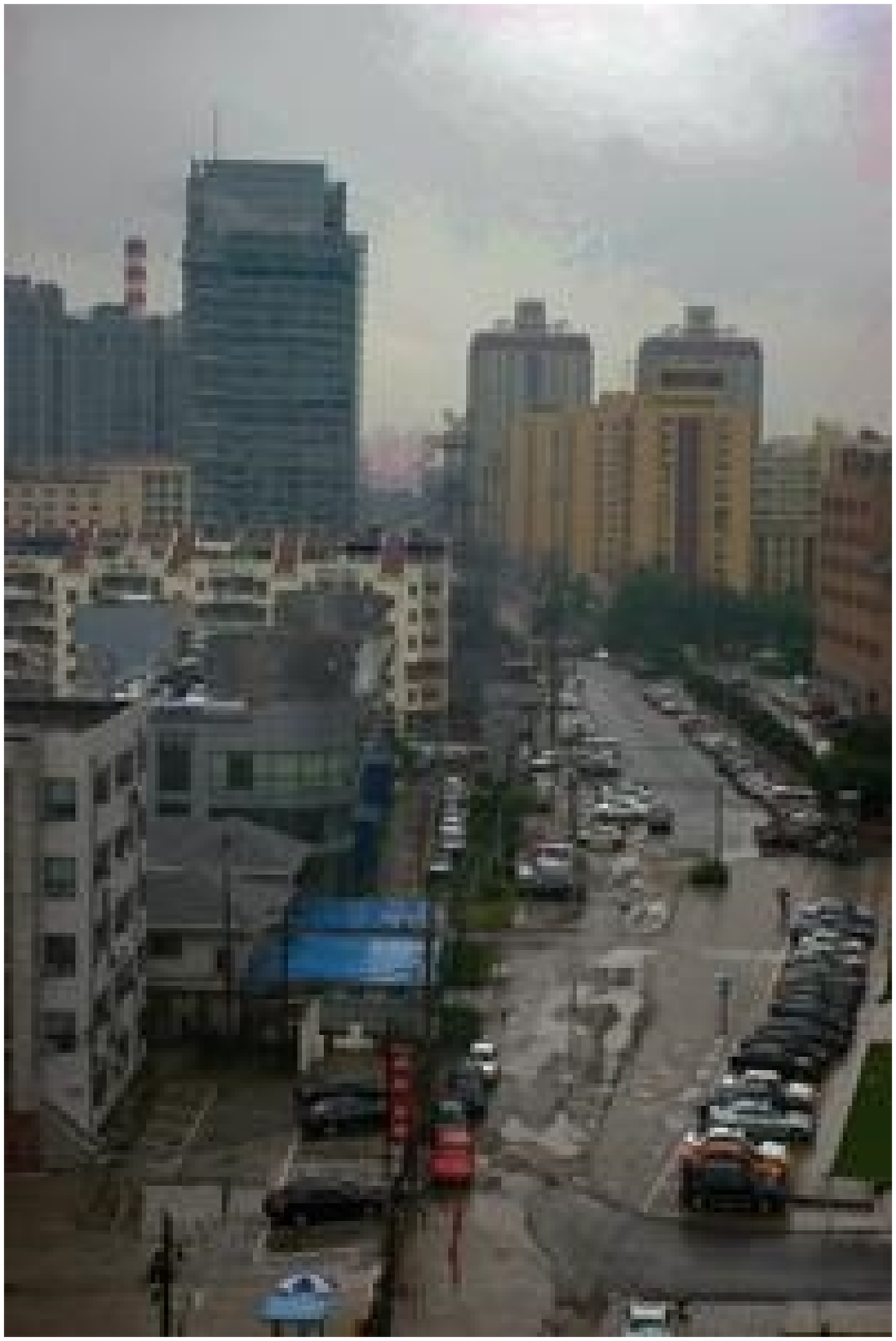}}
\subfigure[NN model]{\includegraphics[scale=0.26]{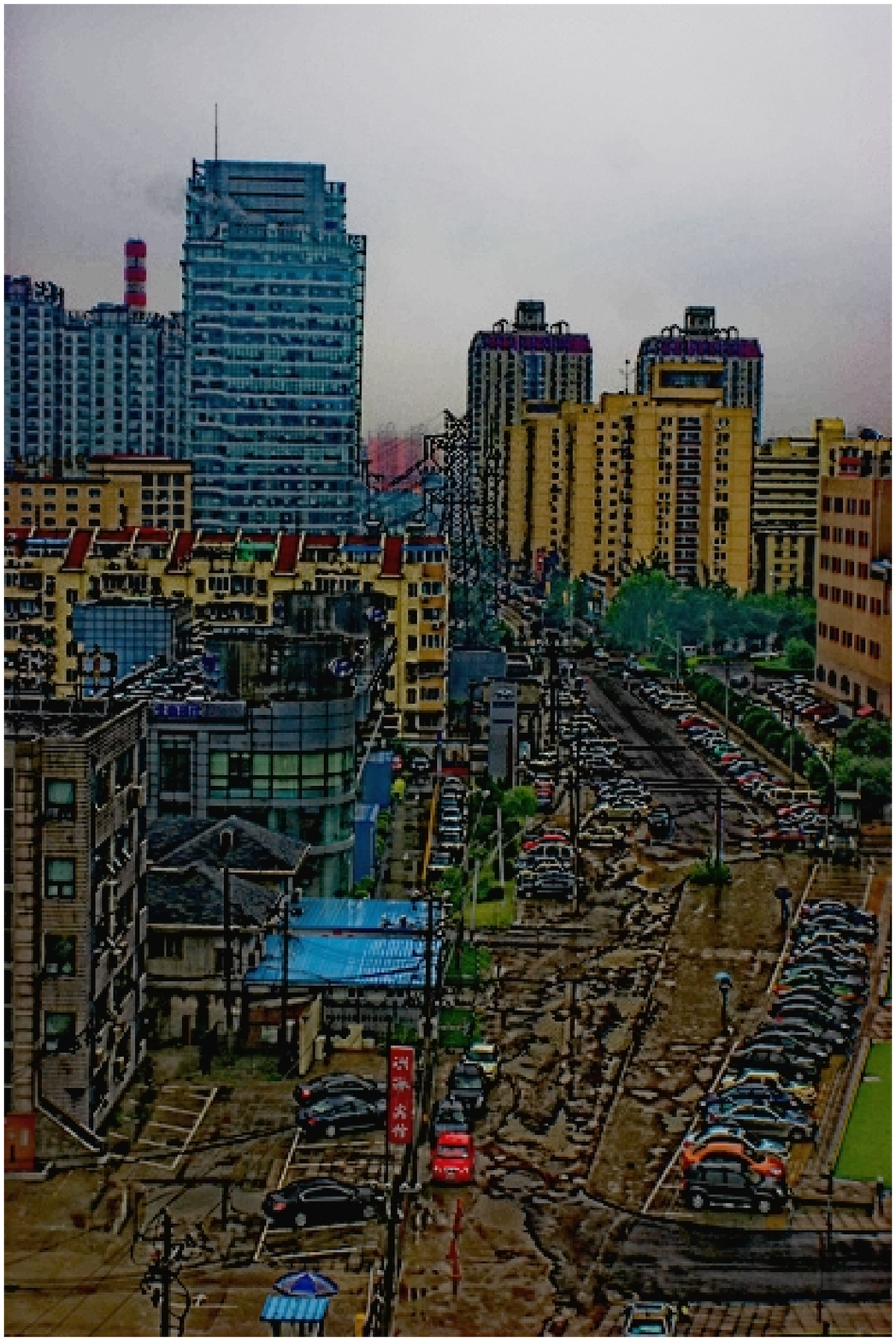}}
\subfigure[GVPB model]{\includegraphics[scale=0.26]{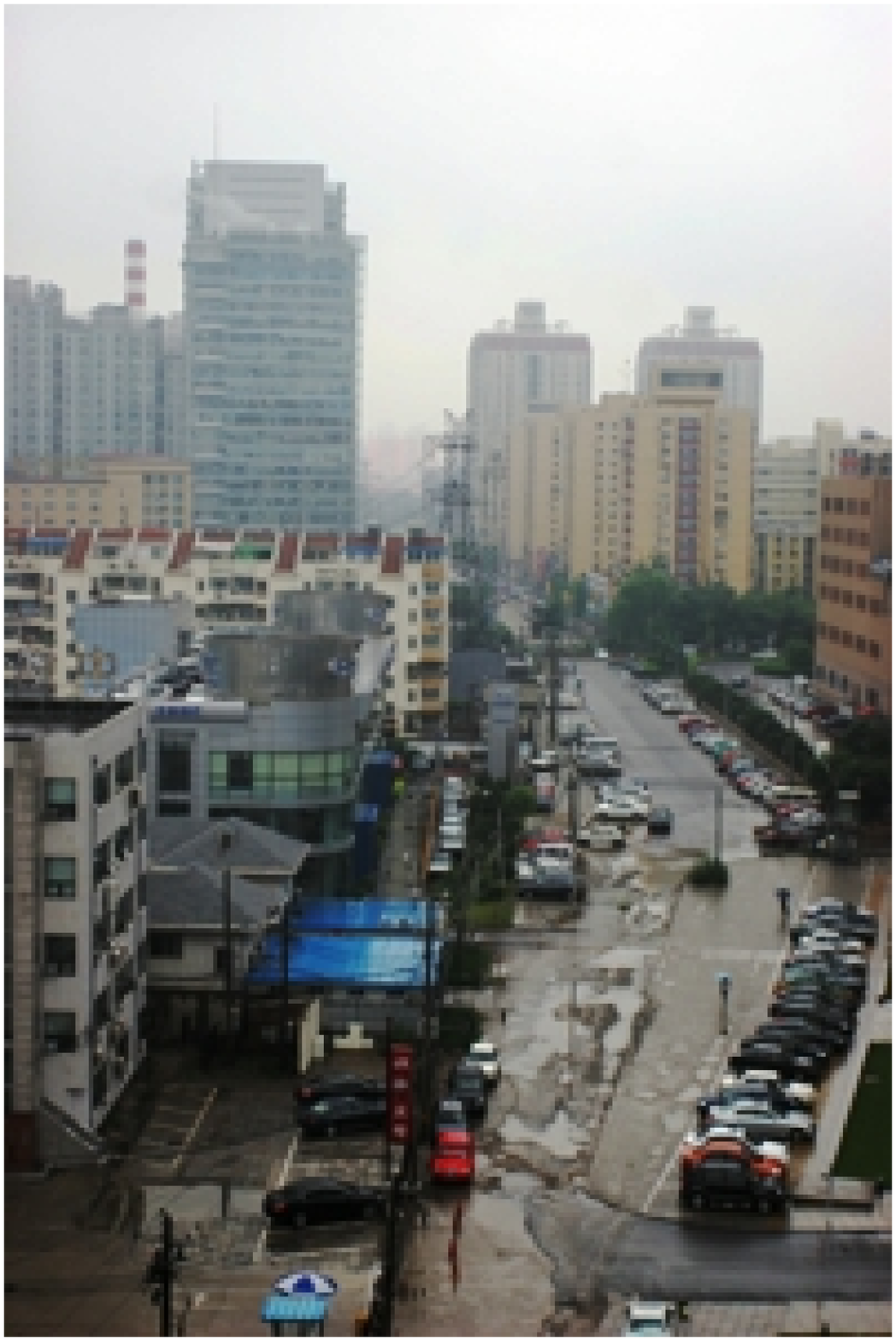}}
\subfigure[Our model]{\includegraphics[scale=0.26]{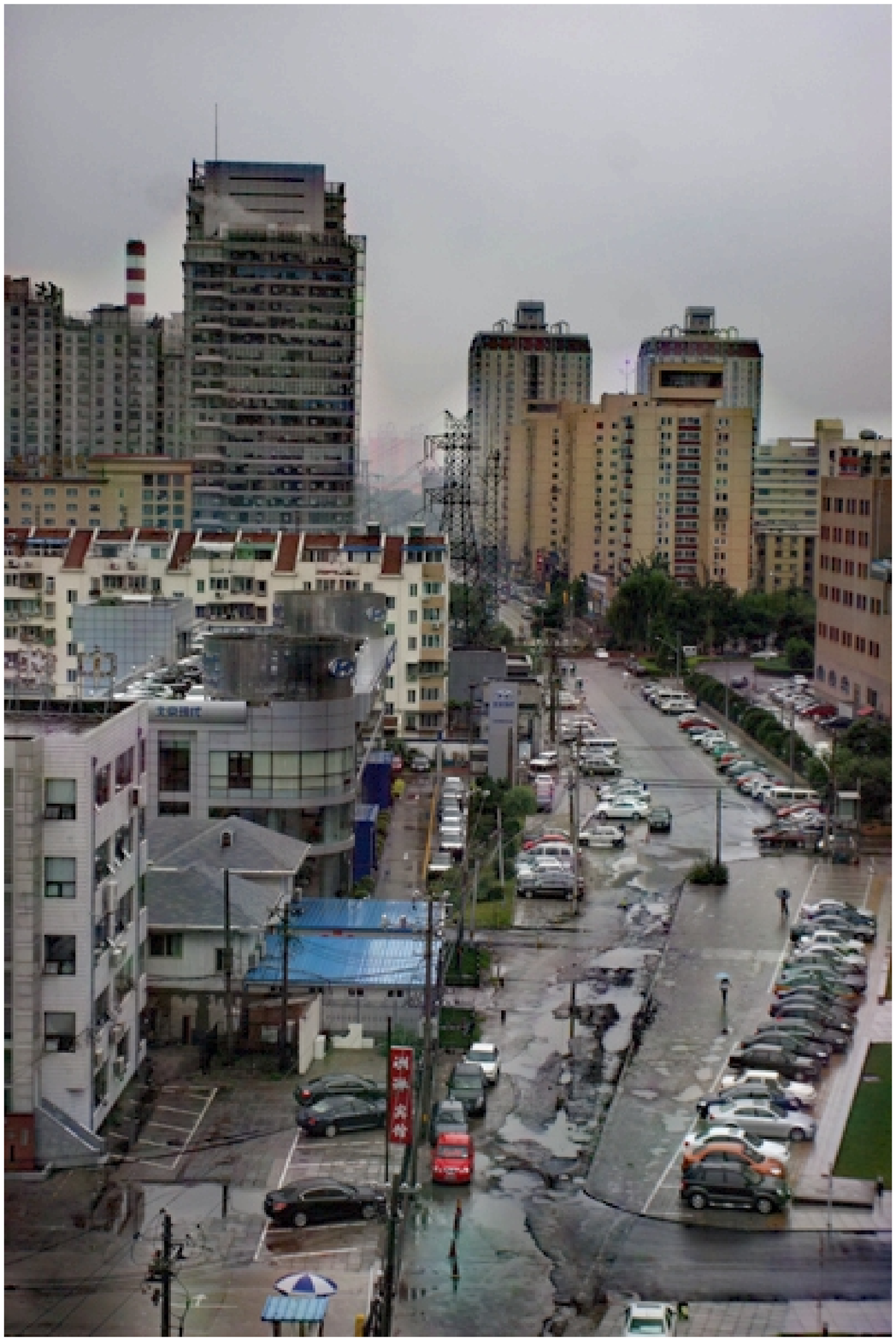}}
\end{center}
\caption{Dehazed images of five models.}
\end{figure*}

\begin{figure*}
\begin{center}
\subfigure[Hazy image]{\includegraphics[scale=0.3]{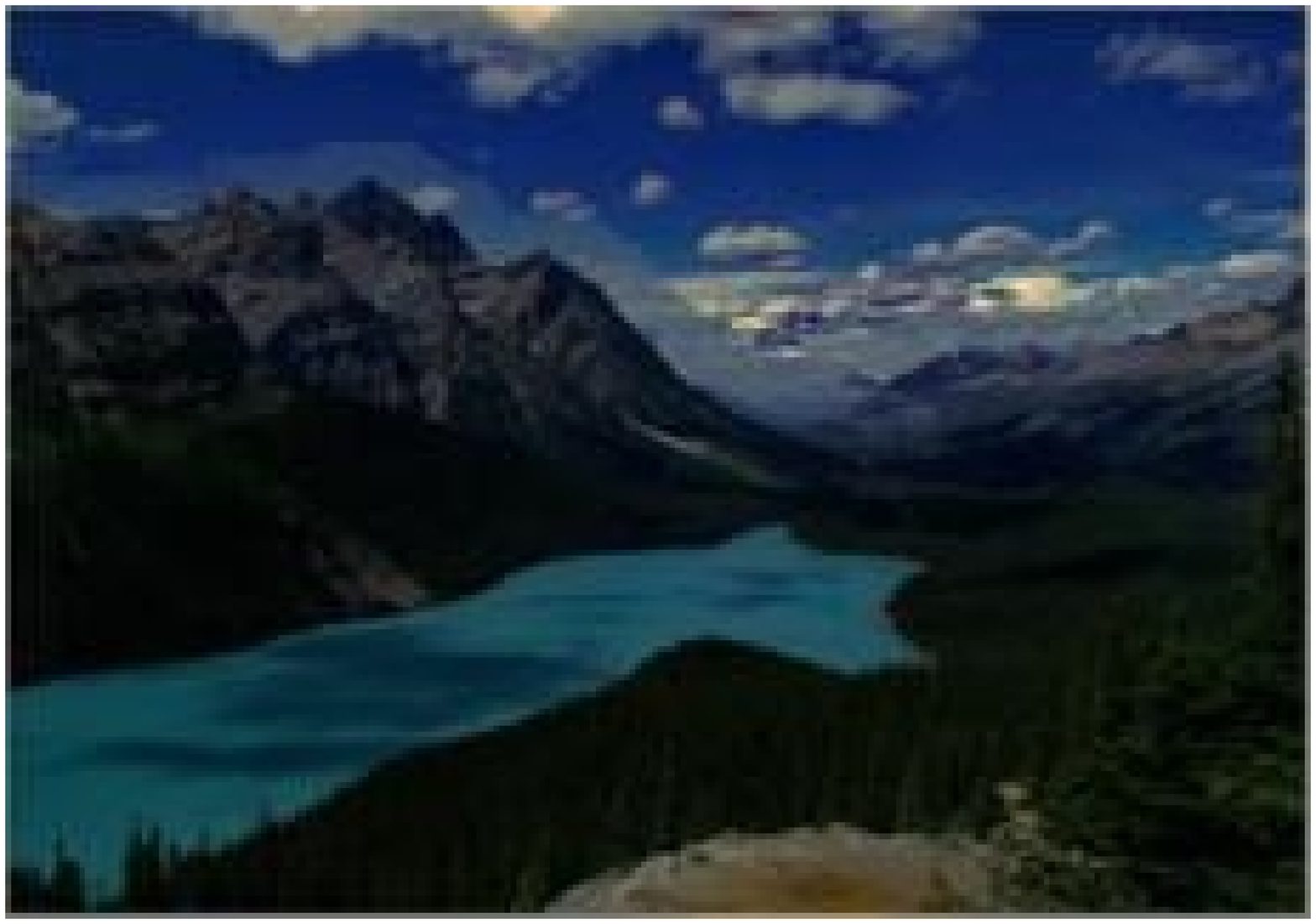}}
\subfigure[TH model]{\includegraphics[scale=0.3]{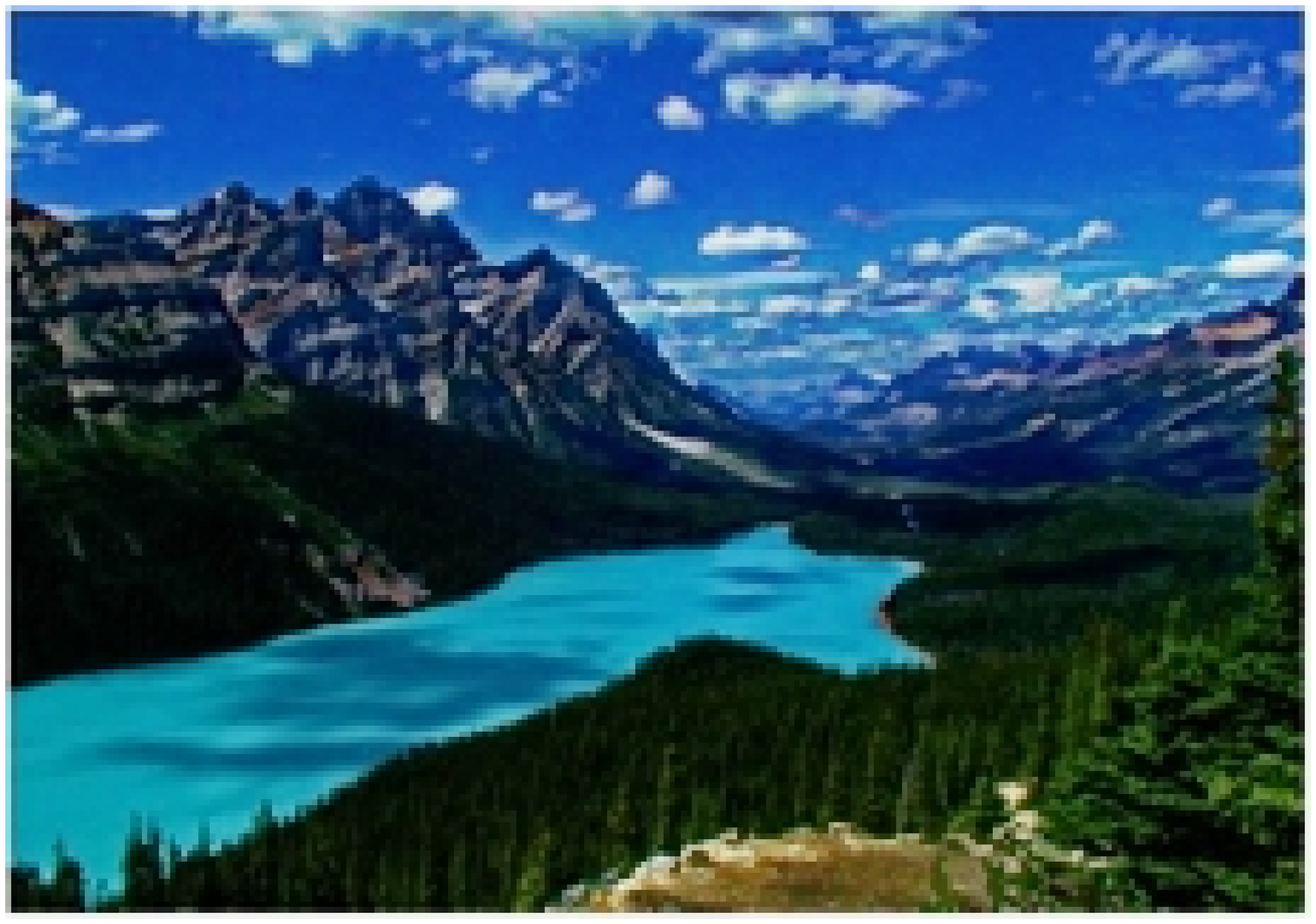}}
\subfigure[HST model]{\includegraphics[scale=0.3]{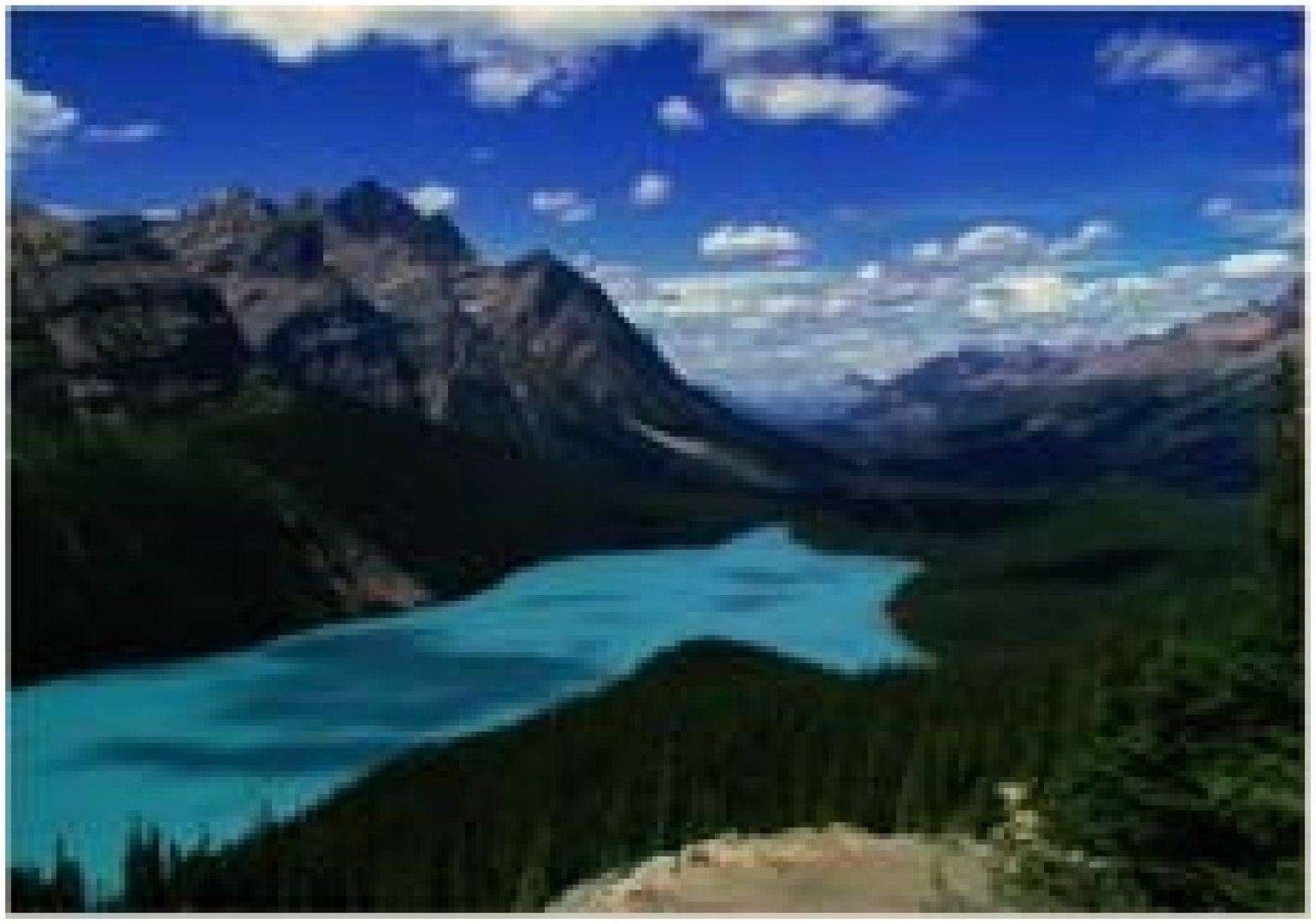}}
\subfigure[NN model]{\includegraphics[scale=0.3]{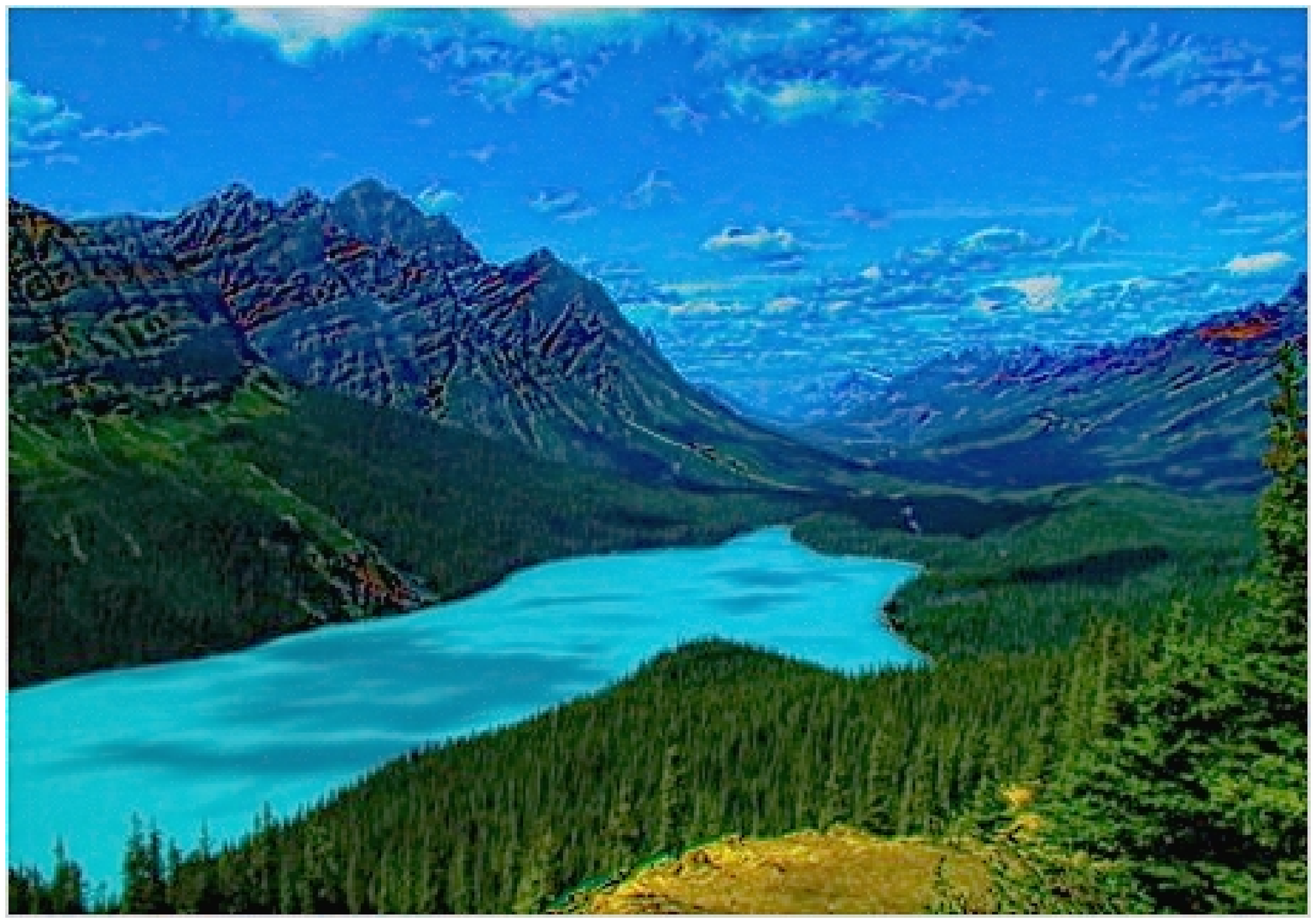}}
\subfigure[GVPB model]{\includegraphics[scale=0.3]{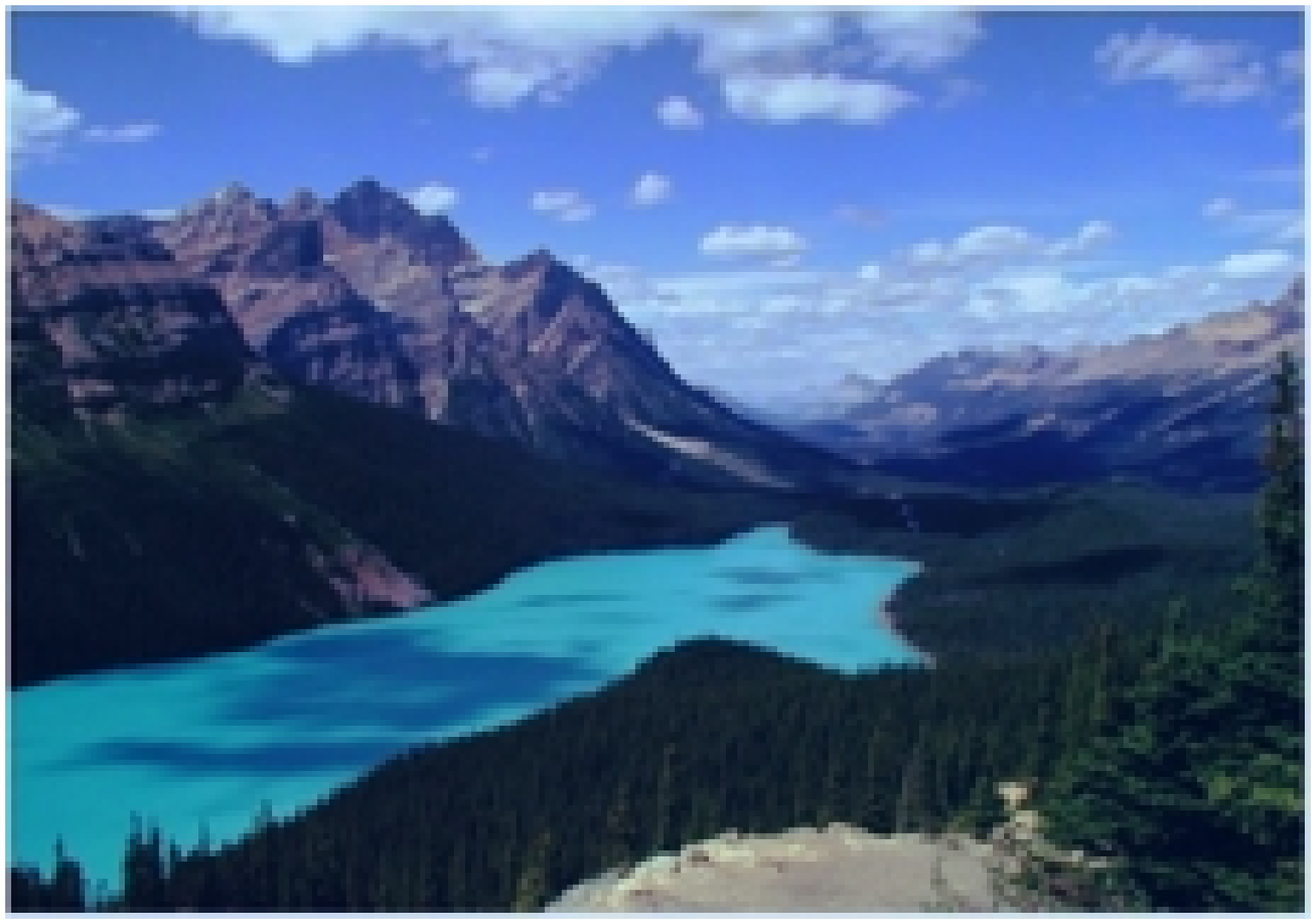}}
\subfigure[Our model]{\includegraphics[scale=0.3]{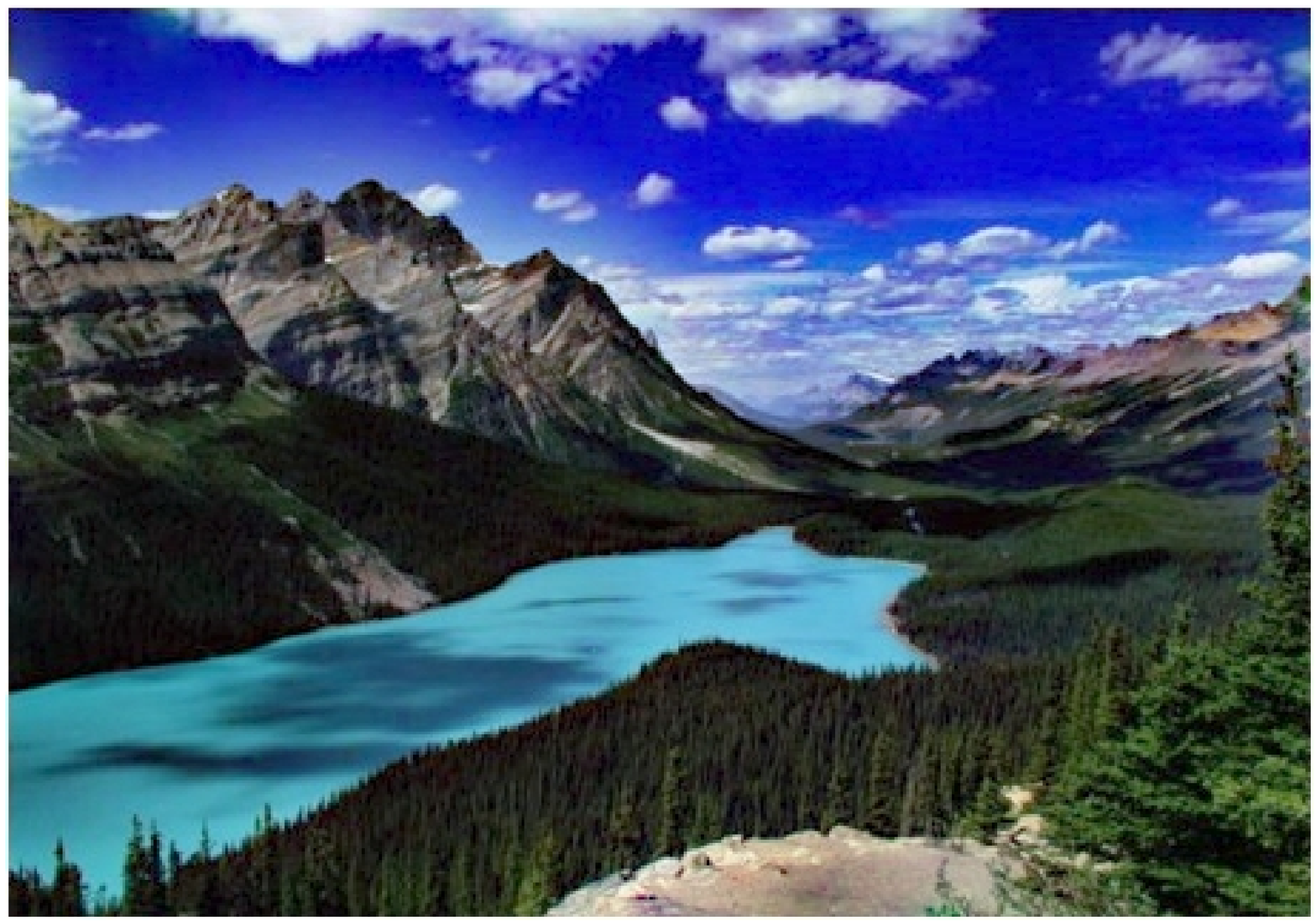}}
\end{center}
\caption{Dehazed images of five models.}
\end{figure*}

\begin{figure*}
\begin{center}
\subfigure[Hazy image]{\includegraphics[scale=0.32]{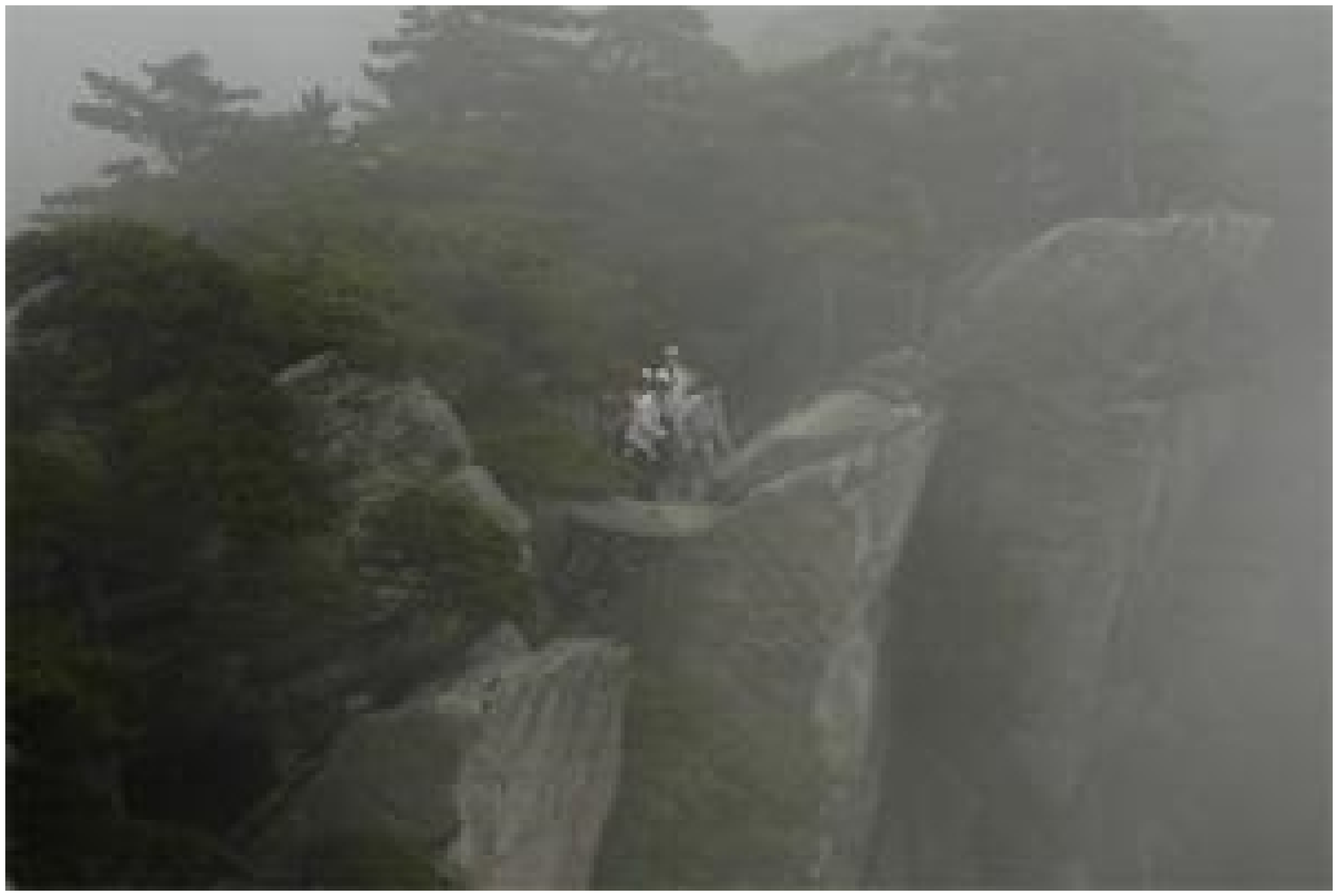}}
\subfigure[TH model]{\includegraphics[scale=0.32]{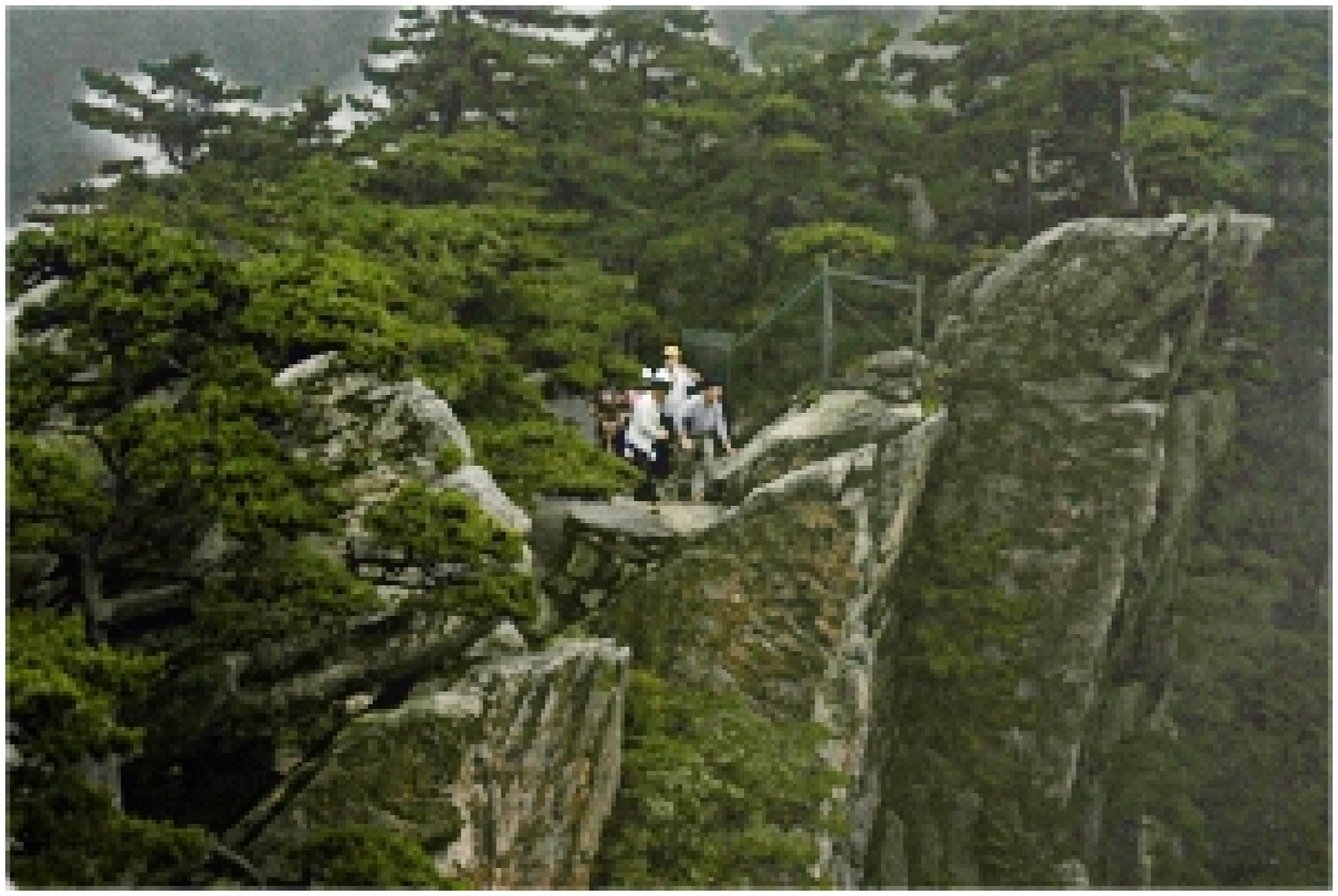}}
\subfigure[HST model]{\includegraphics[scale=0.32]{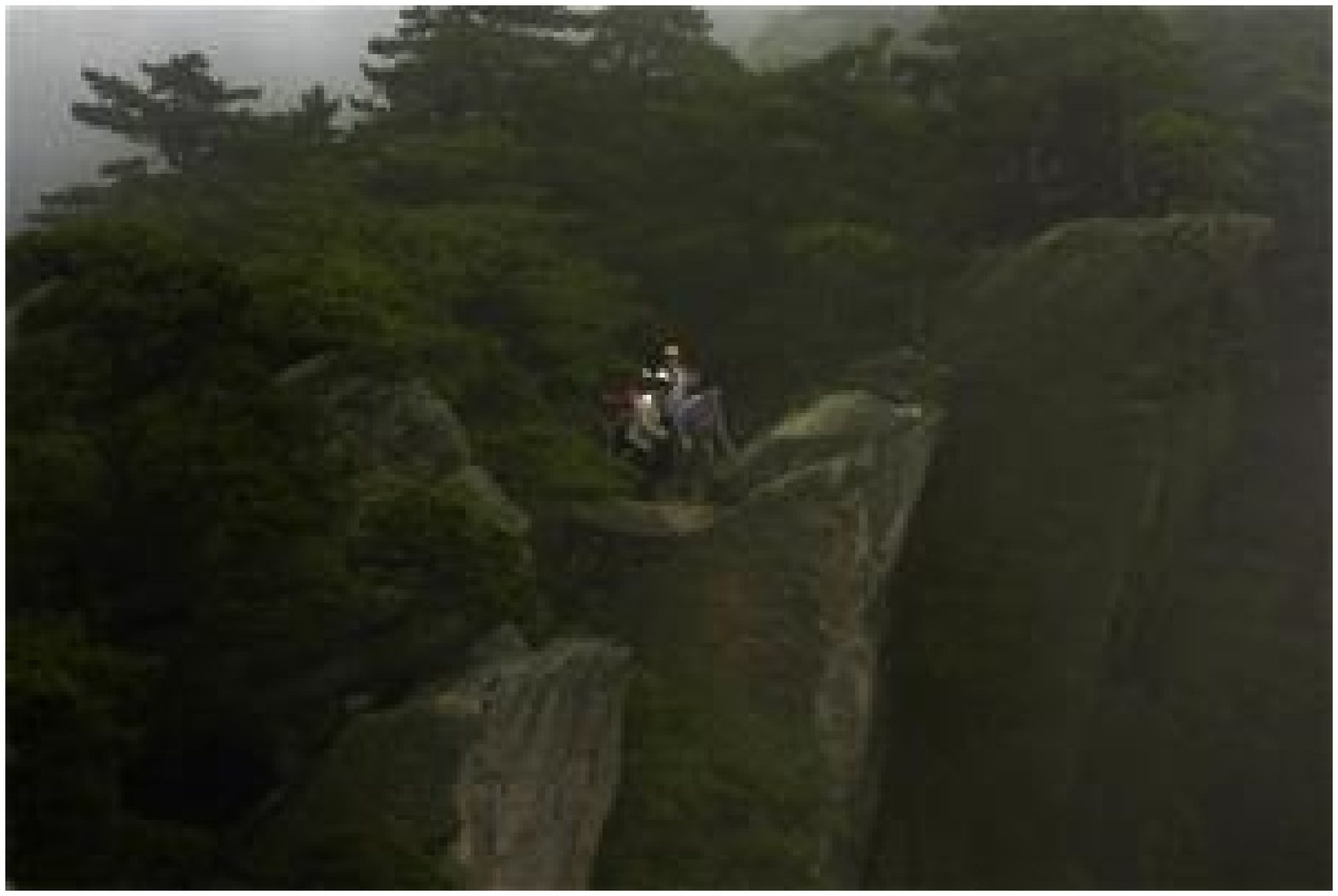}}
\subfigure[NN model]{\includegraphics[scale=0.32]{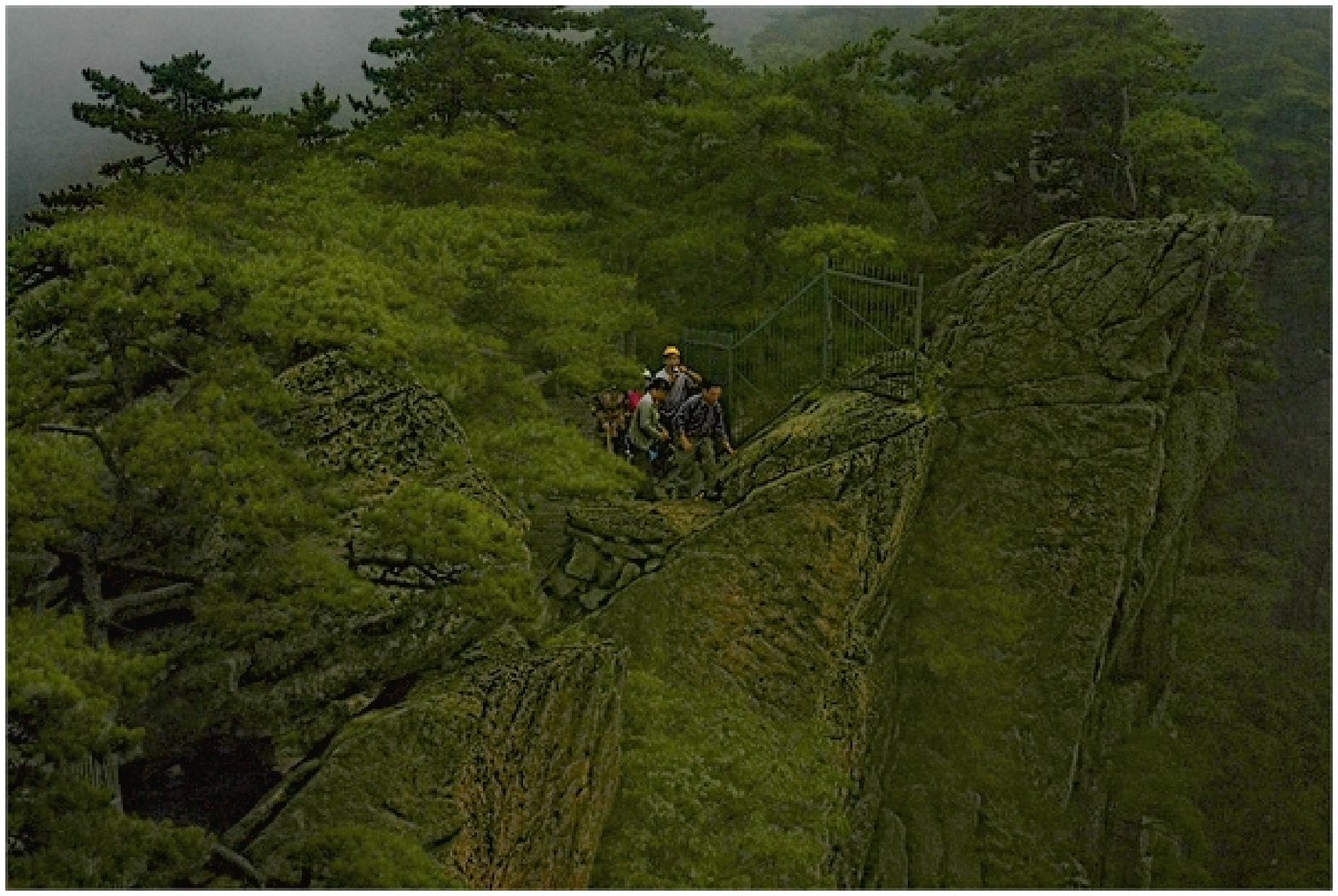}}
\subfigure[GVPB model]{\includegraphics[scale=0.32]{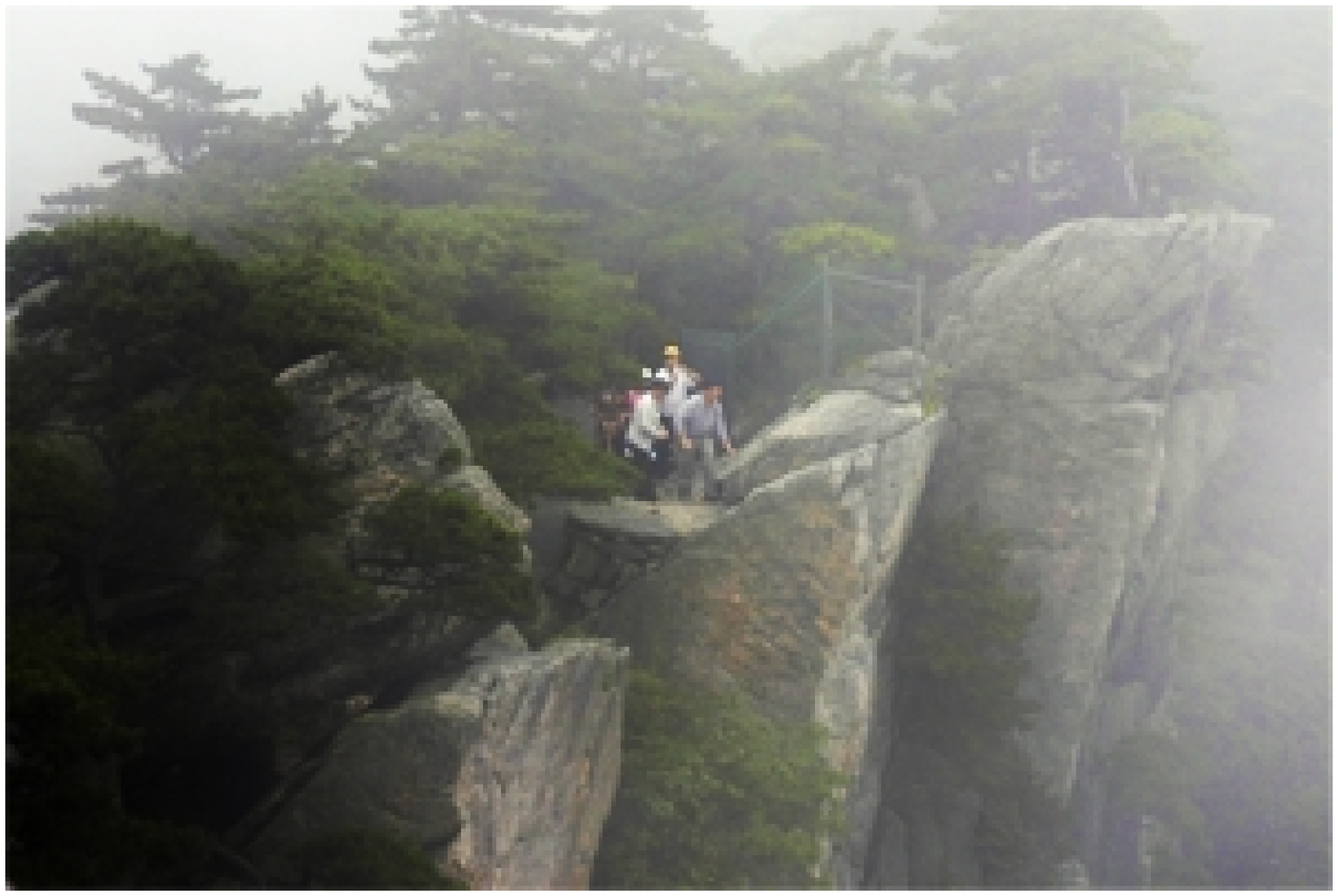}}
\subfigure[Our model]{\includegraphics[scale=0.32]{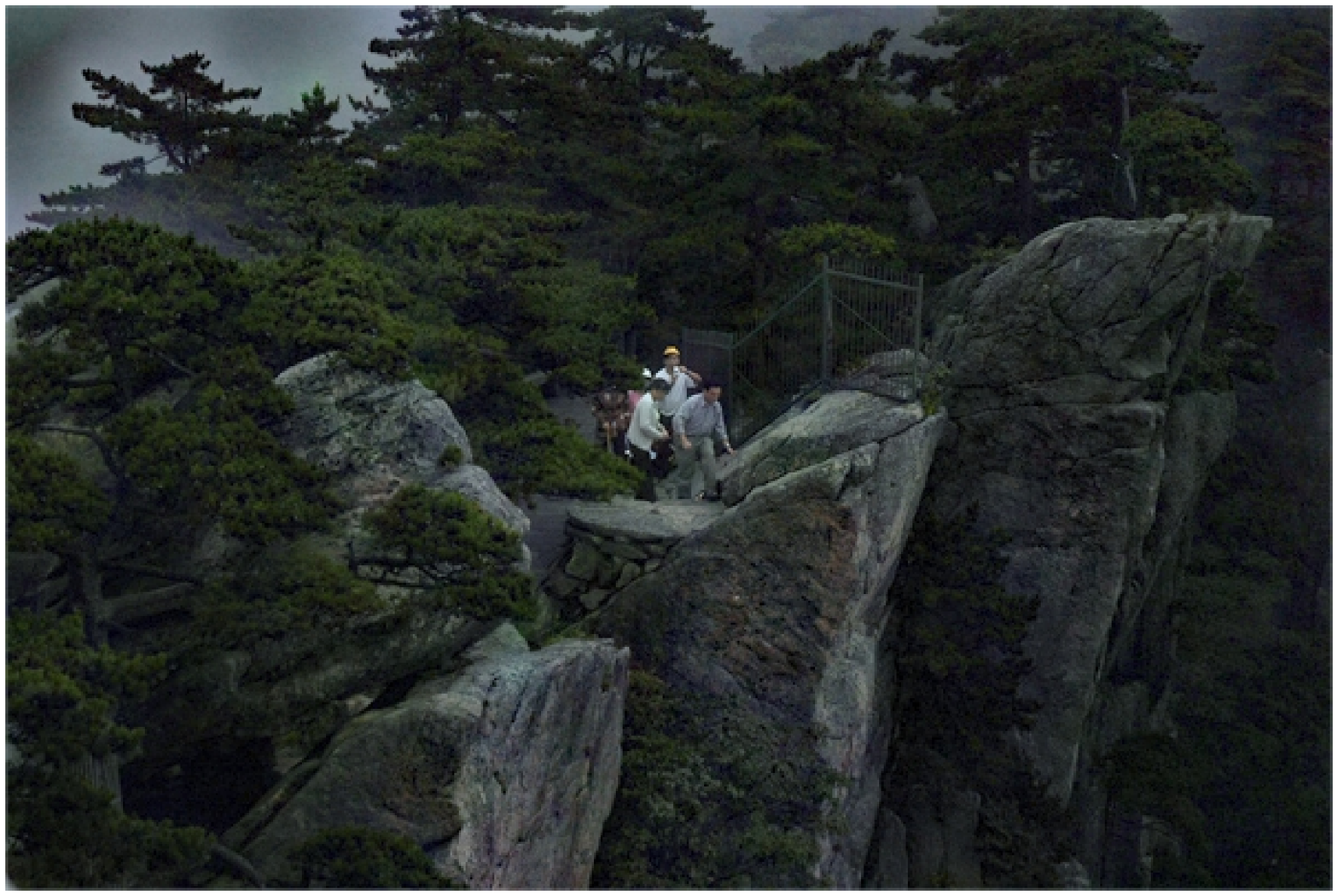}}
\end{center}
\caption{Dehazed images of five models.}
\end{figure*}

\begin{figure*}
\begin{center}
\subfigure[Hazy image]{\includegraphics[scale=0.3]{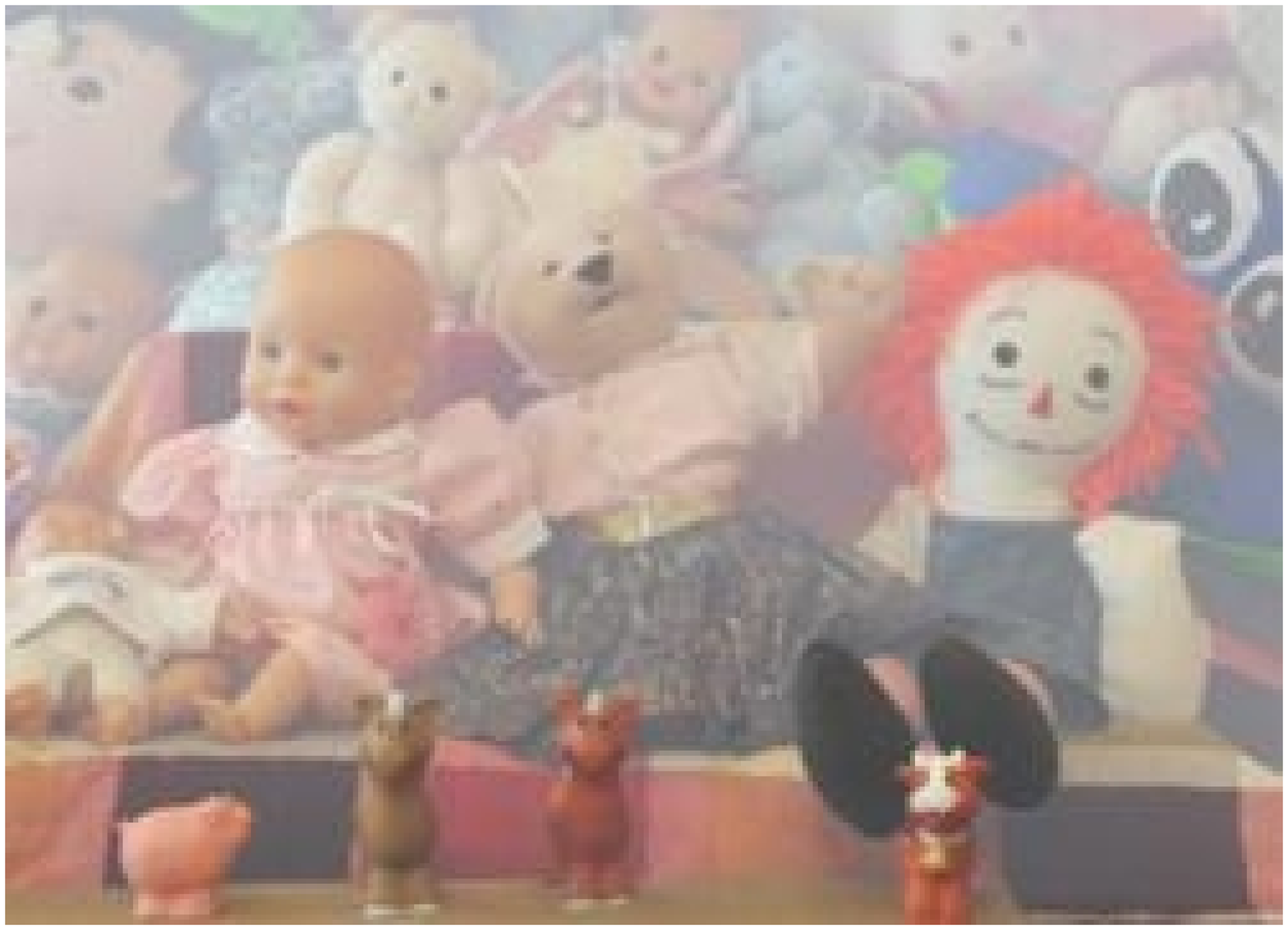}}
\subfigure[TH model]{\includegraphics[scale=0.3]{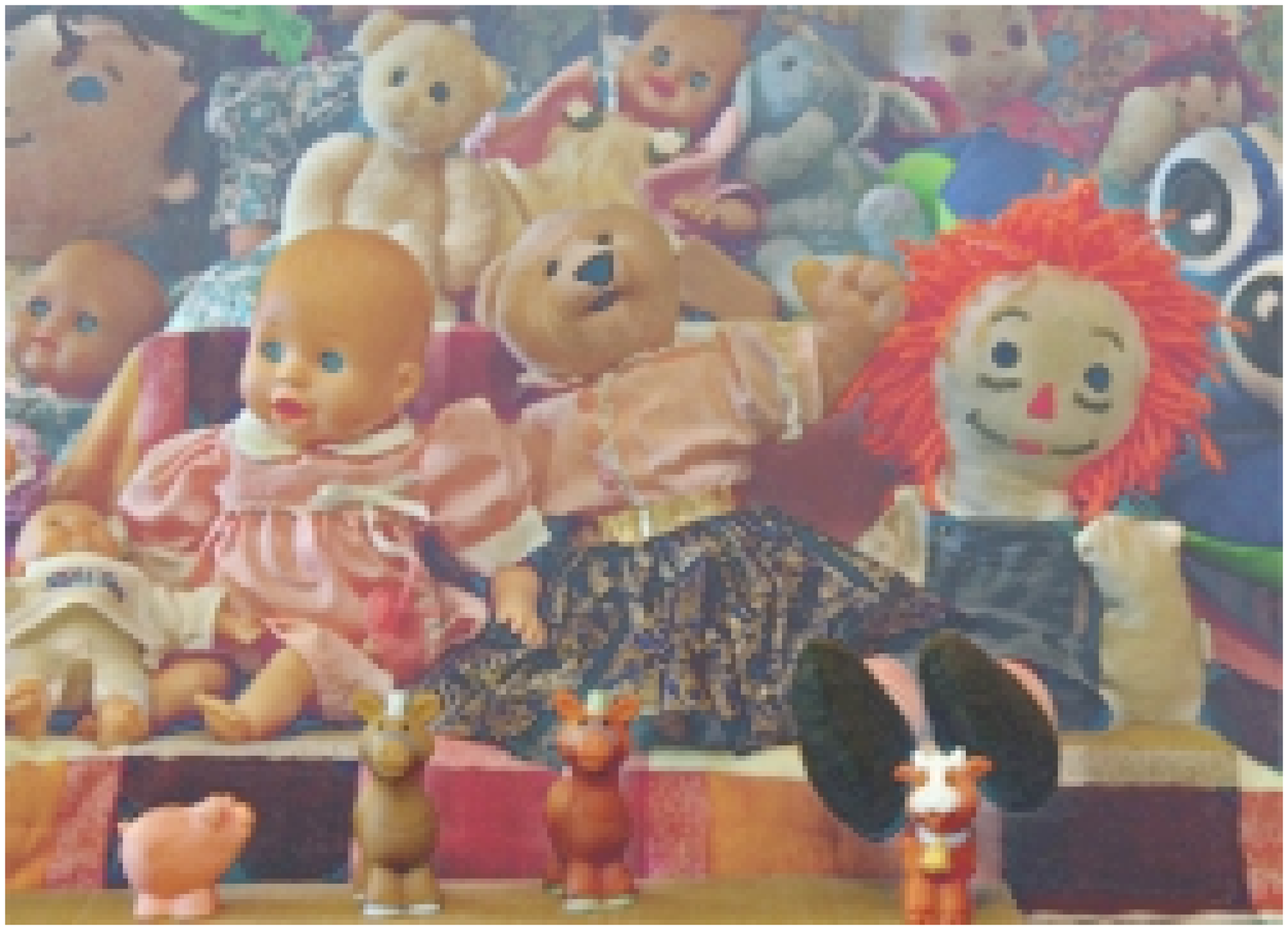}}
\subfigure[HST model]{\includegraphics[scale=0.3]{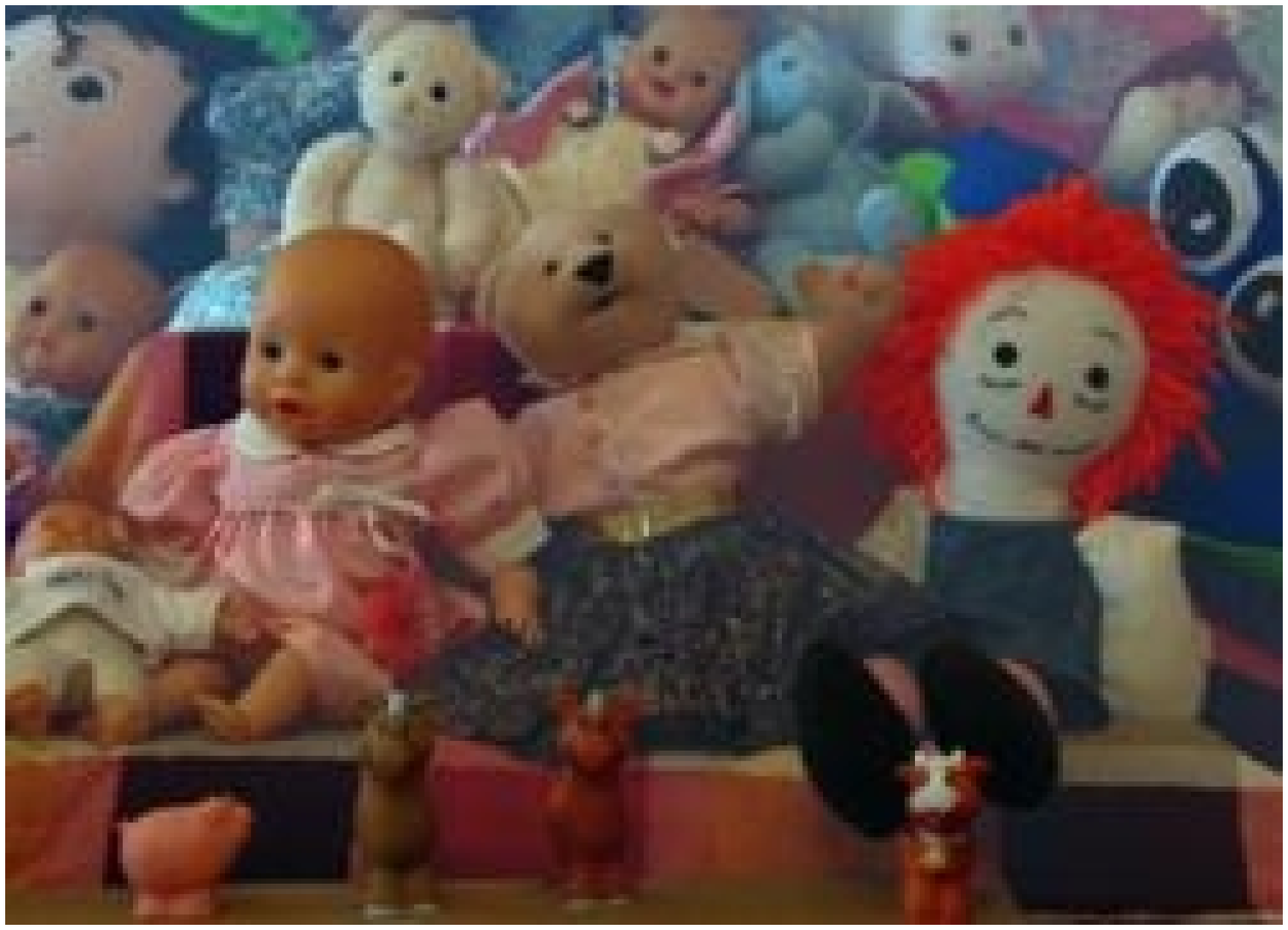}}
\subfigure[NN model]{\includegraphics[scale=0.3]{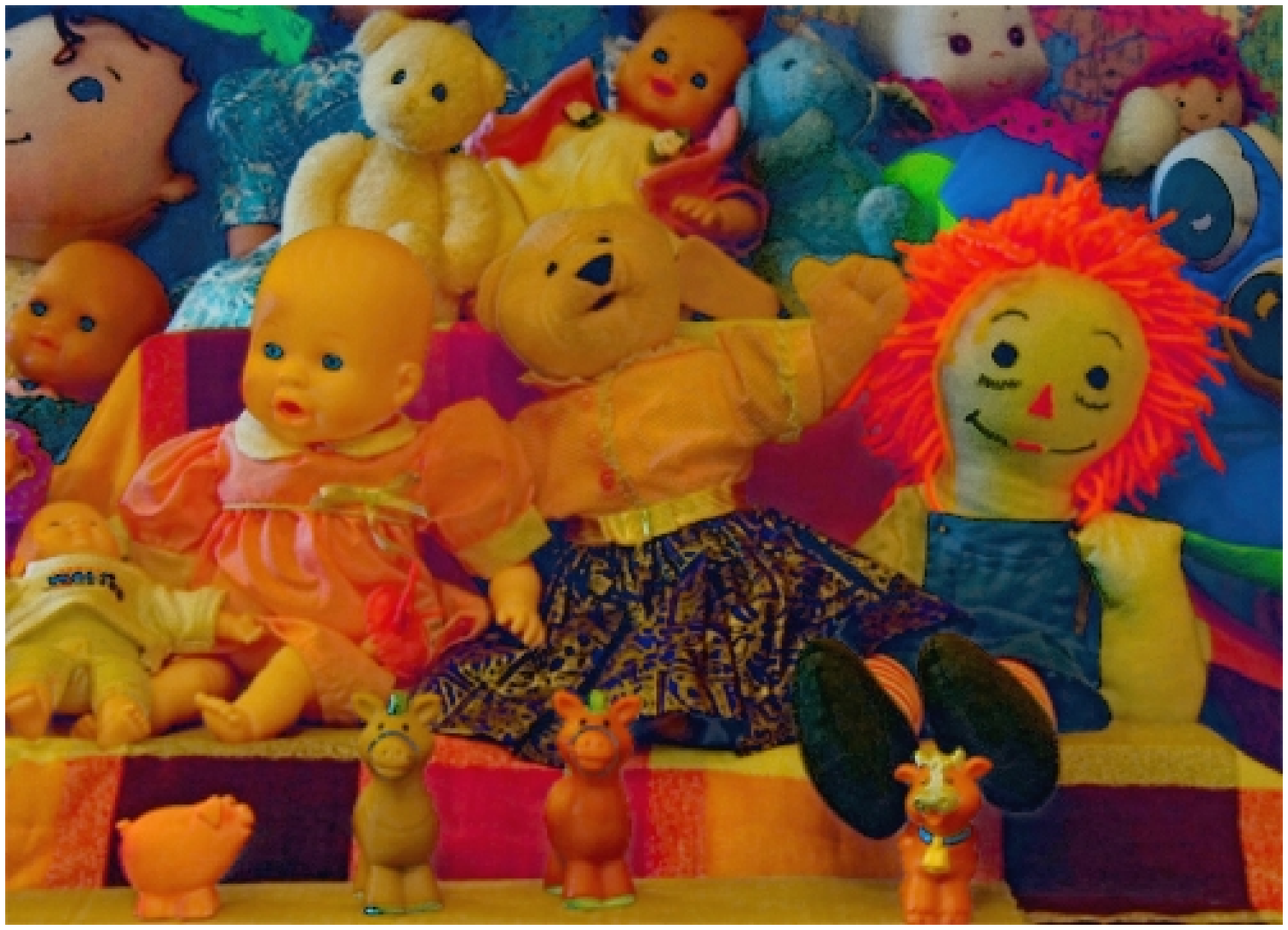}}
\subfigure[GVPB model]{\includegraphics[scale=0.3]{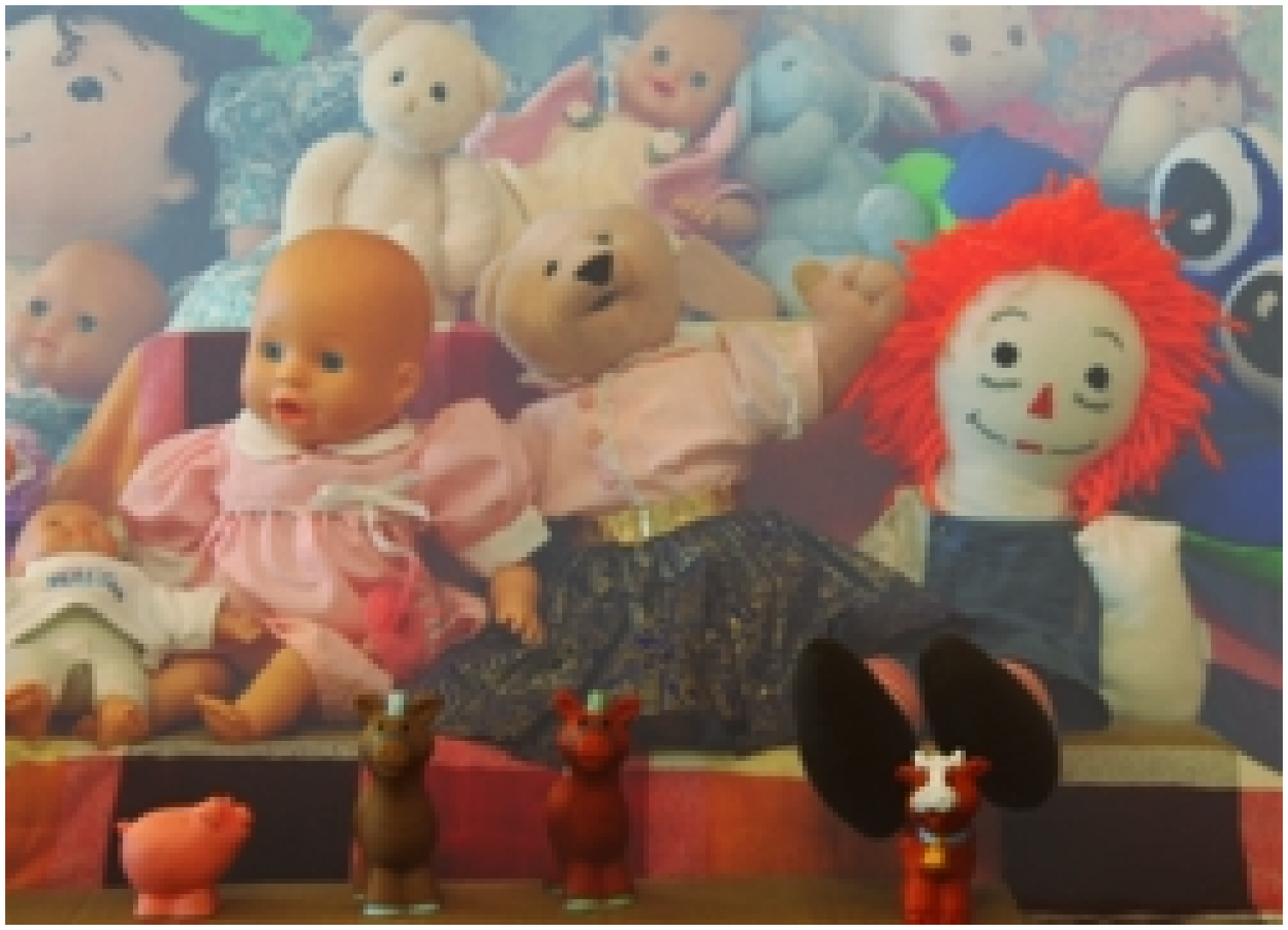}}
\subfigure[Our model]{\includegraphics[scale=0.3]{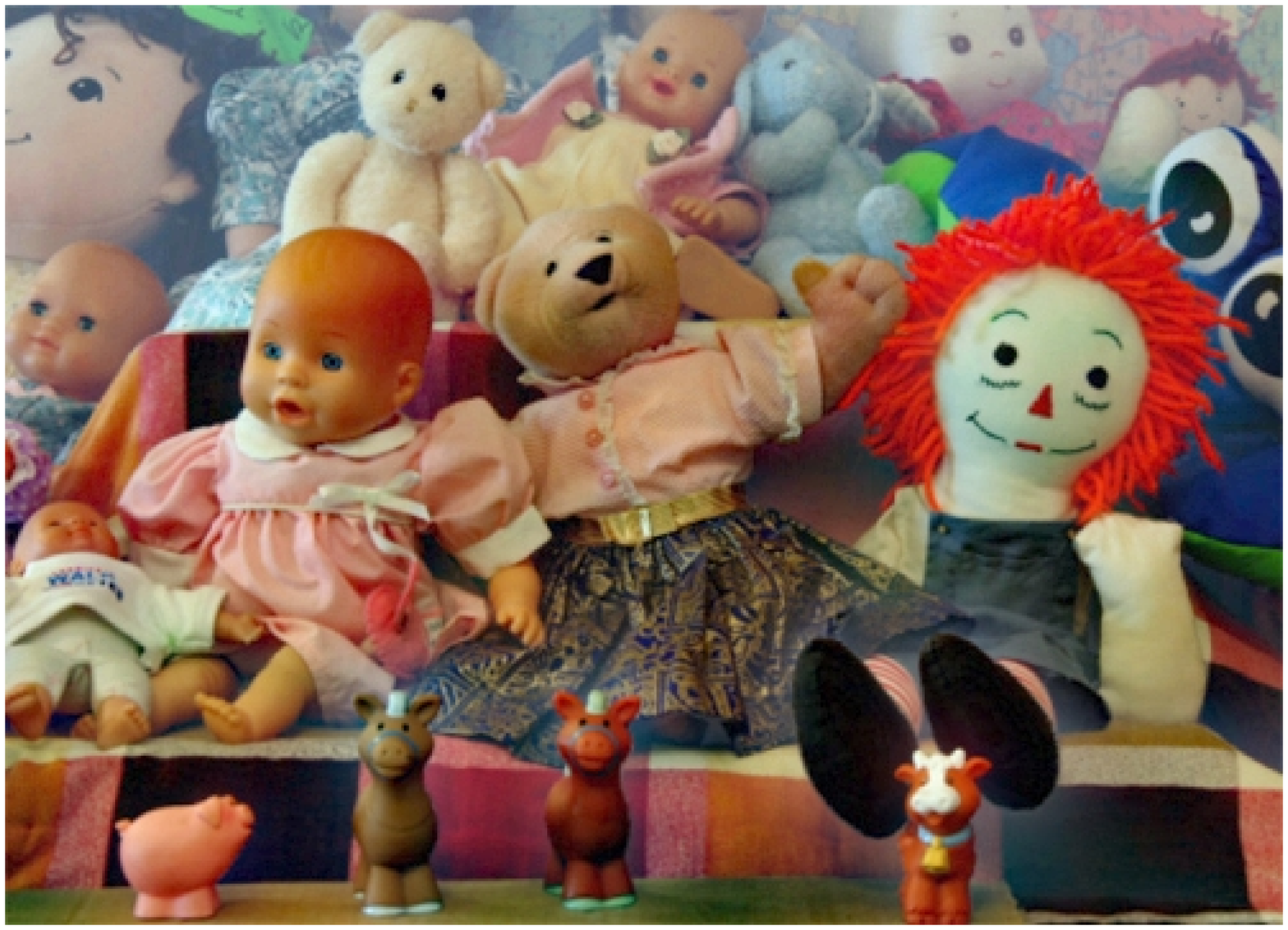}}
\end{center}
\caption{Dehazed images of five models.}
\end{figure*}

\clearpage
\begin{table}\caption{The DIVINE measure.}
\centering  % 表居中
\begin{tabular}{lccccccccc}  % {lccc} 表示各列元素对齐方式，left-l,right-r,center-c
\hline
Figure &TH model        &HST model     &NN model      &GVPB model      &Our model                    \\ \hline  % \hline 在此行下面画一横线
1&69.12939706	&55.81779686	&76.77106426	&27.59361051	&51.70167471\\
2&40.06474732	&50.33357282	&58.90129268	&37.63728444	&28.28020355\\
3&29.5005219	    &31.67365419	&45.48490334	&44.623095	    &31.2784826\\
4&42.361185	    &72.74332432	&69.26949878	&31.99647216	&32.35797595\\
5&34.99726308	&31.71468442	&36.93765104	&58.06464867	&34.36637133\\
6&36.17639358	&42.54461789	&42.01817754	&52.82374224	&32.31507553\\
7&28.25738767	&60.46093524	&60.4186025	    &43.18610658	&36.35281814\\
8&33.92395932	&37.60110402	&44.04815197	&48.93872153	&25.87425821\\
9&37.06484625	&55.50427775	&51.79350658	&26.42666981	&36.64377696\\
10&30.40942193	&43.31822515	&47.75938503	&41.11414384	&31.5831609\\
11&42.59571954	&71.60576038	&62.68818645	&22.42739601	&46.66527958\\
12&34.89638836	&34.32759134	&51.64766252	&35.63436032	&26.64670178\\
13&39.73777801	&45.69397109	&69.62915092	&37.56939792	&25.76558093\\
14&44.01071221	&37.21166345	&55.90721517	&33.26285199	&32.01531235\\
15&37.42343679	&52.63590884	&62.6682146	    &28.92708829	&25.20314234\\
16&37.36820293	&51.08691585	&55.30067248	&46.37173965	&38.32106867\\
17&37.9279738	    &55.30492563	&52.57334629	&29.13674538	&32.11836927\\
18&35.58305551	&35.65267976	&45.32478858	&32.47912919	&31.75197257\\
19&36.22865805	&41.2599423	    &49.21391364	&68.02684495	&40.57380177\\
20&30.8861687	    &102.347466	    &92.92952377	&41.96193037	&33.76610649\\
21&34.18208995	&37.6295786	    &42.56381658	&57.82351567	&26.45331436\\
22&21.01591854	&40.27551157	&43.08234922	&42.08302349	&34.02561442\\
23&50.51538052	&27.63273381	&61.12566397	&55.2424206	    &22.12433037\\
24&37.19092015	&85.19219545	&64.07931102	&26.615661	    &55.31714137\\
25&34.1030265	    &54.10420943	&45.66263991	&49.5616121	    &31.21845141\\
26&33.93844062	&60.76425157	&56.8242781	    &41.99472665    &34.63585255\\
27&26.05382281	&37.24908472	&46.669687	    &40.93523376	&22.93624267\\
28&38.88995823	&94.43701278	&46.3127796	    &44.19758363	&38.66743568\\
29&26.6411135	    &49.3601414	    &47.77859753    &34.81849966	&29.53926097\\
30&19.76311212	&46.78887595	&47.71879282	&48.65305246	&41.65549273\\
31&22.12362886	&47.5608003	    &63.05845156	&39.41038693    &32.53964072\\
32&27.64718521	&63.73911524	&51.03978829	&37.6896322	    &29.40947889\\
33&52.04330222	&77.41422594	&89.29572858	&37.79425201	&28.48594041\\
34&37.80844004	&41.8725081	    &74.21956205	&42.0360692	    &25.22204396\\
35&33.58540955	&33.94907228	&56.72764522	&35.43403771	&24.78363394\\
36&36.88616844	&58.3583864	    &61.6104566	    &40.47316711	&27.74690379\\
37&20.89649786	&39.94838432	&60.51573311	&28.70228926	&31.75203778\\
38&46.97055722	&74.55770339	&77.79395119	&40.05672533	&28.53793788\\
39&39.32233495	&51.09936157	&61.45595098	&34.89836413	&29.95181701\\
40&44.3522329	    &60.50988412	&48.60340461	&39.35458597	&28.23408787\\
41&35.15425653	&63.36065165	&61.28628614	&42.52254652	&36.08010743\\
42&48.01243888	&58.26507293	&80.29961957	&23.74316958	&22.09855608\\
43&31.1217148	    &54.78649132	&58.45596232	&32.89677741	&28.65203484\\
44&50.86545277	&59.54538191	&73.78767393	&33.22594596	&30.74138497\\
45&26.06495856	&54.43095529	&58.19602786	&24.82274303	&26.68134416\\
46&27.32196868	&34.37301228	&48.13772949	&53.17816576	&25.59118287\\
47&60.43678616	&66.0121169	    &66.95232341	&20.23993822	&33.33448687\\
48&36.10163092	&69.32864735	&70.28794576	&39.73012258	&25.50152289\\\hline
Average&36.6157&53.1538&58.2256&39.0903&31.7812\\\hline
\end{tabular}
\end{table}

\end{document}